\documentclass[runningheads]{llncs}

\usepackage{eccv}

\usepackage{eccvabbrv}

\usepackage{graphicx}
\usepackage{booktabs}

\usepackage[accsupp]{axessibility}  

\usepackage{multirow}

\usepackage{hyperref}

\usepackage{orcidlink}

\definecolor{justin}{RGB}{0,0, 0} 

\usepackage{pifont}
\newcommand{\cmark}{\ding{51}} 
\newcommand{\xmark}{\ding{55}} 

\begin{document}

\title{	
High-speed Imaging through Turbulence with Event-based Light Fields} 


\author{Yu-Hsiang Huang\inst{1}
\and
Levi Burner
\inst{1}
\and 
Sachin Shah
\inst{1}
\and 
Ziyuan Qu
\inst{2}
\and 
Adithya Pediredla
\inst{2}
\and
Christopher A.~Metzler\inst{1}
}


\authorrunning{Y.-H.~Huang et al.}

\institute{University of Maryland, College Park MD 20742, USA \and
Dartmouth College, Hanover NH 03755, USA 
\vspace{1em}
\url{https://justhowww.github.io/lf-ev-turb-project-page} 
}

\maketitle

\begin{abstract}
This work introduces and demonstrates the first system capable of imaging fast-moving extended non-rigid objects through strong atmospheric turbulence at high frame rate.
Event cameras are a novel sensing architecture capable of estimating high-speed imagery at thousands of frames per second. However, on their own, event cameras are unable to disambiguate scene motion from turbulence. 
In this work, we overcome this limitation using event-based light field cameras: By simultaneously capturing multiple views of a scene, event-based light field cameras and machine learning-based reconstruction algorithms are able to 
disambiguate motion-induced dynamics, which produce events that are strongly correlated across views, from turbulence-induced dynamics, which produce events that are weakly correlated across view.
Tabletop experiments demonstrate event-based light field can overcome strong turbulence while imaging high-speed objects traveling at up to 16,000 pixels per second.

\end{abstract}

\begin{figure}[h]
    \centering
    \includegraphics[width=0.9\linewidth]{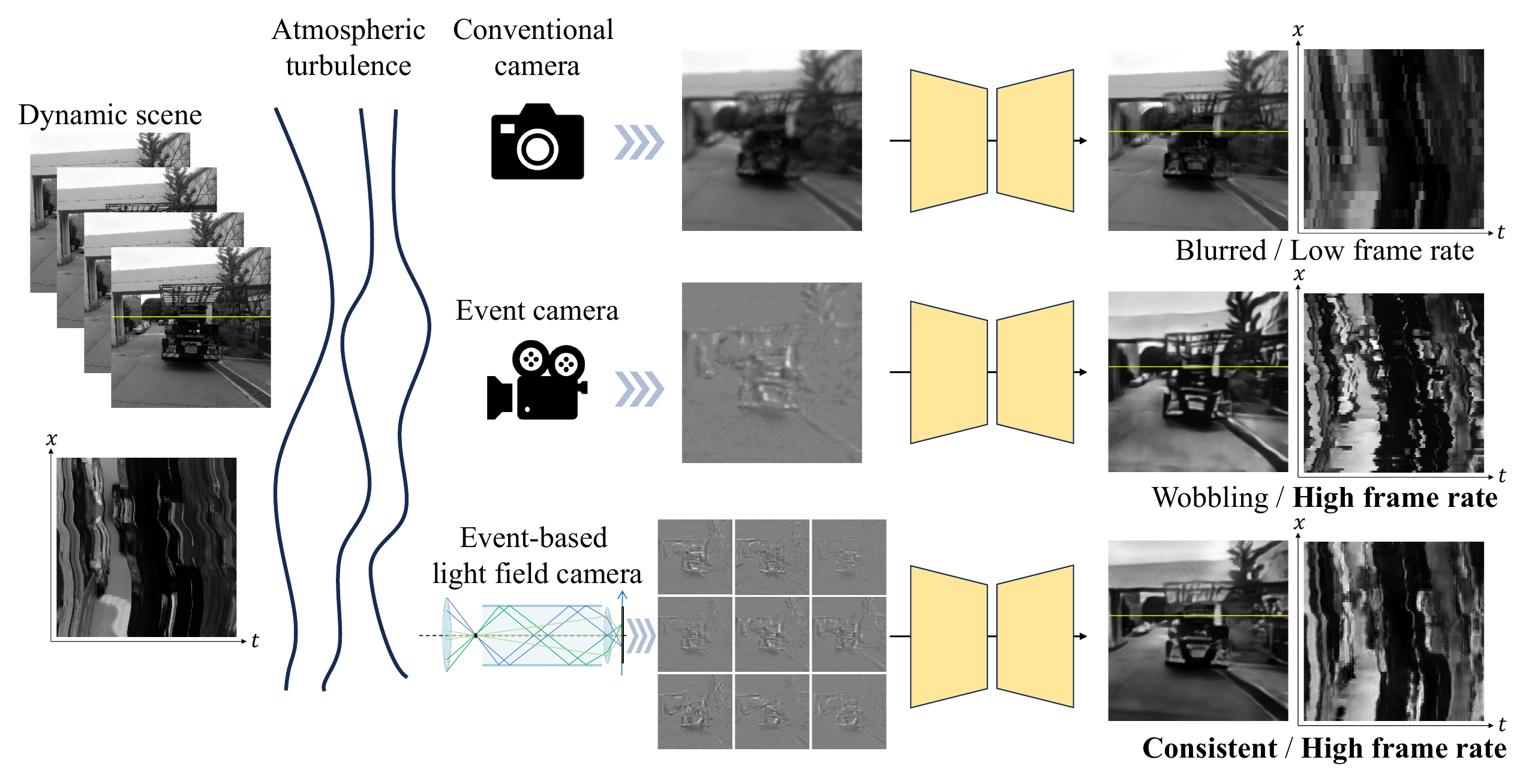}
    \caption{\textbf{Method overview.} We introduce an event-based light field camera to image through turbulence at high speed. The event camera avoids motion blur by detecting brightness changes at microsecond time resolution. The light field offers cross-view constraints on the scene.}
    \label{fig:placeholder}
\end{figure}
\section{Introduction}
\label{sec:intro}

Long-range imaging through the atmosphere is the primary goal of terrestrial astronomy, remote sensing, surveillance, and other applications. 
Atmospheric turbulence represents the central impediment to this goal. 
Regardless of sensor resolution, aperture size, platform stability, or cost, atmospheric turbulence significantly degrades image fidelity by introducing blur, geometric distortion, and wander.

Traditionally, turbulence has been mitigated with hardware-based solutions, such as adaptive optics (AO)~\cite{hampson2021adaptive}, or software-based solutions, such as lucky imaging~\cite{fried1978probabilitylucky} or, more recently, deep learning~\cite{datum, mambatm, tmt}.
However, almost all existing solutions share the same fundamental constraint: 
they rely upon conventional image sensors which integrate light over a fixed exposure window, irrecoverably blending any scene dynamics or turbulence fluctuations that occur over a time scale shorter than the exposure time. 
In this work, we overcome this limitation with event cameras.

Event cameras mitigate motion blur by measuring brightness change asynchronously at microsecond timescales, making them a natural candidate for recovering high-speed dynamics.
Several lines of work have used frames and events to image through turbulence~\cite{boehrer2019eventatm,boehrer2021turbulence,evturb,egtm}. However, they have relied upon restrictive rigid-body assumptions about scene motion~\cite{boehrer2019eventatm,boehrer2021turbulence} or have restricted the reconstructions to the rate of the frame-based camera~\cite{evturb,egtm}, thereby giving up event-camera's central advantage.


Event-only imaging of dynamic scenes through turbulence is nontrivial. 
If the scene were static and all events were due to turbulence, one may be able to exploit these events to help reconstruct the scene~\cite{he2024microsaccade}. 
If the turbulence was static and the scene was dynamic, one could use existing event-to-video frameworks to reconstruct a (distorted) high-speed video~\cite{e2vid, firenet, e2vid+}. 
However, when both the turbulence and the scene are dynamic, both will produce (largely indistinguishable) events, thereby invalidating the assumptions that existing methods rely upon.


In this work, we use an event-based light field camera (Event Fields)~\cite{qu2025eventfield} to overcome this ambiguity.
Light fields provide multiple sub-apertures simultaneously, each corrupted by a different turbulence realization, while scene dynamics remain largely consistent across all views.
Thus, each light field view provides a constraint on the singular scene image, and inconsistent motion across views can be attributed to turbulence.
We exploit this structure (implicitly) using the recurrent encoder-decoder architecture E2VID~\cite{e2vid} with stacked multi-view event inputs and train it to reconstruct clean, high-speed video through turbulence.

In summary, our contributions are:
\begin{itemize}
    \item The first event-only method for high-speed imaging through turbulence, breaking the frame-rate barrier of conventional cameras.
    \item Demonstration that light field optics provide sufficient information for removing turbulence while preserving scene motion.  
    \item Real (in-lab) and simulated experiments demonstrating our method significantly outperforms single-view reconstruction.
\end{itemize}

\section{Related works}

\subsection{Light Field Imaging}
Light field imaging acquires multiple simultaneous views of a scene from distinct viewpoints, encoding both spatial and angular information of light~\cite{lightfieldsurvey2016, lightfieldsurvey2017}. 
Camera arrays achieve this by positioning multiple independent sensors at spatially separated viewpoints~\cite{yang2002camarray}. 
While conceptually straightforward, they require precise inter-sensor synchronization and their physical bulk limits deployment in constrained platforms.
Single-shot alternatives using microlens arrays overcome the synchronization problem by imaging all sub-apertures onto a single sensor~\cite{adelson1992singlelenslightfield, ng2005lightfieldcamera}.
However, the dense lenslet grid demands precise optical alignment and manufacturing tolerances. 
Kaleidoscope-based designs avoid both drawbacks~\cite{manakov2013reconfigurable, qu2025eventfield}.
Planar mirrors placed at the intermediate image plane fold the optical path, routing light from distinct regions of the aperture to separate areas of the sensor and producing multiple virtual viewpoints without additional optics per channel.
This construction requires no microlens fabrication and eliminates inter-camera synchronization entirely, accordingly we adopt it here.


\subsection{Imaging Through Aberrations with Light Fields} 
Ng and Hanrahan first demonstrated that light fields could be used to correct for optical aberrations back in 2006~\cite{ng2006lfaberration}. 
Light fields have since been used to image through turbulence on multiple occasions~\cite{Wu2016lfthruturbulece,loktev2011specklelfturbulence,wu2019comparisonlfturbulence,Wu2022integratedsensor}. 
Intuitively, by capturing multiple perspectives of the same scene at the exact same time, these cameras create valuable spatial redundancy that can be used to remove turbulence and other aberrations.
These systems have relied on conventional image sensors. While a few recent works have combined light fields with event cameras to perform high-speed 3D imaging~\cite{guo2024eventlfm,qu2025eventfield}, ours is the first work to use event-based light field cameras to image through turbulence.

\subsection{Software-based Imaging through Turbulence}

Classical turbulence mitigation algorithms often use spatial registration and temporal fusion~\cite{Mao2020tmstanley, caliskan2014tmwithopticalflow, 2008tmspline} to recover imagery. Accordingly, these methods generally depend on static references or distortion-free patches and fail in dynamic scenes or strong turbulence. Learning-based methods relax these requirements: self-supervised and test-time optimization use priors such as lucky images~\cite{fried1978probabilitylucky}, blind degradation estimation~\cite{nert, Li_2021_unsupervised}, or pretrained mitigation models~\cite{feng2023turbugan,cai2024temporally} and adapt without paired data, but are iterative and ill-suited to high-speed use; meanwhile physics-based simulators~\cite{turbulence_p2s} enable supervised learning with fast inference via feed-forward sequence-to-sequence~\cite{tmt} or recurrent~\cite{datum, ge2025eventmamba} networks. All of these approaches remain limited by conventional sensors’ fixed exposure, which conflates scene dynamics and turbulence at high frame rates.



Event cameras, with microsecond-level temporal resolution, are a natural candidate for resolving this bottleneck. An early work, used an event camera to identify lucky regions in conventional frames \cite{boehrer2019eventatm,boehrer2021turbulence}.
Two recent works explore this direction with machine learning: EvTurb~\cite{evturb} employs events to guide single-frame restoration, while EGTM~\cite{egtm} leverages event temporal resolution to compute adaptive fusion weights for conventional frames.
In both, the final image formation depends on the integration time of the conventional sensor, leaving output constrained to standard frame rates.

A fundamental bottleneck thus persists across all existing methods: event cameras alone cannot distinguish scene-induced events from turbulence-induced events.
We propose an event-only approach that leverages light field imaging to resolve this ambiguity and achieve high-speed imaging through turbulence.

\subsection{Hardware-based Imaging through Turbulence}
AO, wherein one measures and then physically compensates for optical aberrations using deformable mirror or spatial light modulator, is the primary hardware-based approach for imaging through turbulence~\cite{hampson2021adaptive}. 
The majority of AO systems rely upon wavefront sensors~\cite{watnik2018wavefront,guo2024direct,chimitt2025wavefront}, which can be accelerated with event cameras~\cite{kong2020shack,ziemann2024learning,wang2024neuromorphic,grose2024convolutional}. 
However, with a few notable exceptions~\cite{yeminy2021guidestar,feng2023neuws,xie2024wavemo,guo2024direct,jiang2026guidestar,pellizzari2026speckle}, existing AO systems rely upon guidestars---bright points in the scene that can be used for calibration---which greatly limits their flexibility. Ours is the only guidestar-free turbulence mitigation approach that can operate at kHz.

\subsection{Events to Video}
Reconstructing intensity frames from event streams has seen rapid evolution.
Early methods~\cite{e2vid,firenet,e2vid+} employ recurrent U-Net structures with ConvLSTM or ConvGRU modules to accumulate temporal state across asynchronous events, enabling continuous video reconstruction.
HyperE2VID~\cite{ercan2024hypere2vid} extends this paradigm by introducing a hypernetwork that dynamically predicts convolutional kernel sizes, improving adaptability to varying event densities.
To better capture long-range temporal dependencies, ETNet~\cite{etnet} adopts transformer architectures, achieving improved reconstruction quality at the cost of increased computational overhead.
More recently, EventMamba~\cite{ge2025eventmamba} leverages state-space models to achieve efficient long-range modeling with linear complexity. A fundamental assumption underlies all these methods: events arise exclusively from scene dynamics or camera motion.  

Under atmospheric turbulence, events no longer arise from scene motion alone.
With a single camera, both turbulence and scene dynamics introduce changes in intensity, and retraining these networks on turbulent data does not resolve this fundamental ambiguity.
We overcome this limitation with a light field event camera that simultaneously captures multiple views of the scene.
Events caused by scene dynamics are strongly correlated across views, while events caused by turbulence are weakly correlated, providing the cross-view structure a network can exploit to disentangle the two.

\section{Methods}
\begin{figure}[!t]
    \centering
    \includegraphics[width=\linewidth]{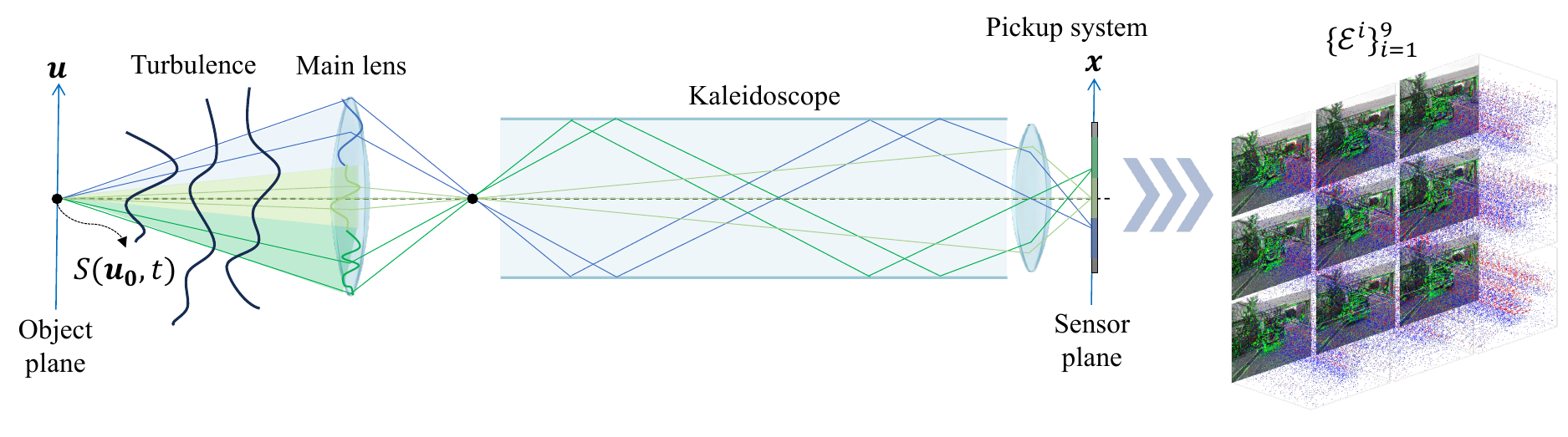}
    \caption{\textbf{Image model for event-based light field camera under turbulence.} A light field camera captures $N$ simultaneous sub-aperture views, each observing the same scene through an independent turbulence realization, which injects different turbulence-induced fluctuations across views while leaving the scene contribution identical.}
    \label{fig:imaging_model}
\end{figure}
\subsection{Modeling Turbulent Event Light Fields}


Our imaging model is illustrated by \autoref{fig:imaging_model}. A detailed description follows. The intensity at the sensor $I(\mathbf{x},t)$ is given by the incoherent imaging integral under an anisoplanatic point spread function (PSF):
\begin{equation}
    I(\mathbf{x},t) = \int_{\Omega_s} |h_\mathbf{u}(\mathbf{x},t)|^2\, S(\mathbf{u},t)\, d\mathbf{u},
    \label{eq:image_formation}
\end{equation}
where $S(\mathbf{u},t)$ is the scene radiance and $h_\mathbf{u}(\mathbf{x},t)$ is the complex coherent PSF which incorporates the phase errors introduced by turbulence along the optical path~\cite{goodman2005introduction}. 
The accumulated phase error leads to the geometric warp and spatially varying blur observed at the sensor, as dictated by $h_\mathbf{u}(\mathbf{x},t)$~\cite{chan2023turbulence_textbook}.

This forward model is bilinear in $S$ and $|h|^2$: scene motion (changes in $S$) and turbulence (changes in $h$) are multiplicatively coupled, making their individual contributions ambiguous in the observed intensity $I$.
Thus, recovering both from a single observation constitutes an ill-posed blind inverse problem.

Multiple views mitigate this ambiguity. If $N$ observations share a scene $S$ through separate turbulence realizations $\{h^{(i)}\}$, the system accumulates cross-view constraints on the common $S$:
\begin{equation}
    I^{(i)}(\mathbf{x},t) = \int_{\Omega_s} |h^{(i)}_\mathbf{u}(\mathbf{x},t)|^2\, S(\mathbf{u},t)\, d\mathbf{u}, \quad i = 1,\ldots,N.
    \label{eq:multiview}
\end{equation}

$S$ can be assumed to be the same in each view, because at long range parallax is negligible.
However, turbulence affects each view differently. Rays entering each sub-aperture traverse a spatially distinct atmospheric column, imparting a unique instantaneous phase error and associated PSF $h^{(i)}_\mathbf{u}(t)$.

We obtain $N$ views with a light field event camera, with the goal of high speed imaging through turbulence. Following the standard event camera model~\cite{e2vid}, the intensity of each view is passed through a non-linear logarithmic threshold: an event $(\mathbf{x}, t, p) \in \mathcal{E}^{(i)}$ is emitted whenever the absolute log-intensity change $\Delta L^{(i)}(\mathbf{x},t) = |\log I^{(i)}(\mathbf{x},t) - \log I^{(i)}(\mathbf{x},t{-}\Delta t)|$ exceeds a threshold $C$, with polarity $p_i = \operatorname{sgn}(\Delta L^{(i)})$.

\begin{figure}[t]
\centering
  \includegraphics[width=1\linewidth]{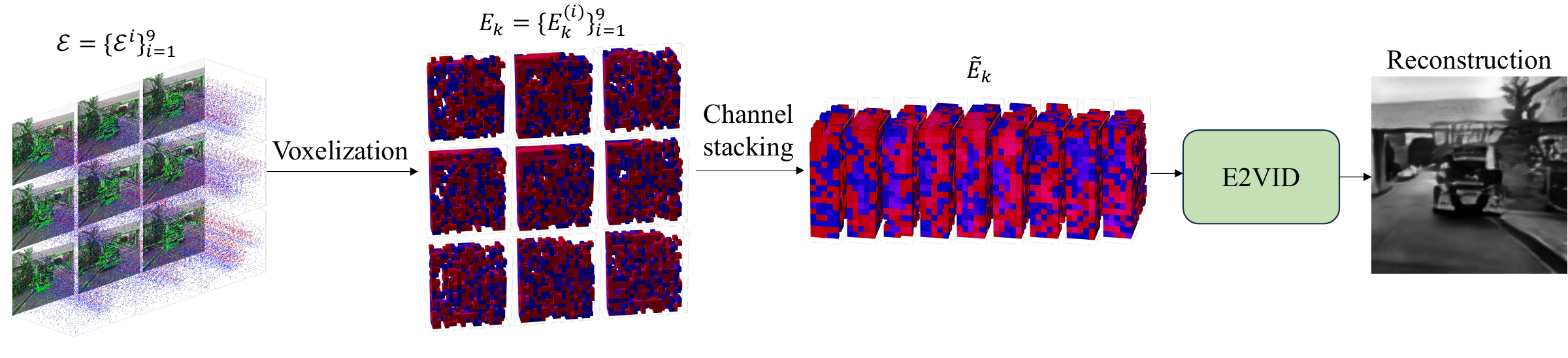}
\caption{\textbf{Network architecture.} Each sub-aperture event stream is converted to a voxel grid $E_k^{(i)} \in \mathbb{R}^{B \times H \times W}$; the $N$ grids are stacked along the channel dimension to form the joint input $\tilde{E}_k \in \mathbb{R}^{NB \times H \times W}$. A recurrent convolutional encoder-decoder processes the stacked input window by window and outputs a reconstructed frame $\hat{V}_k$.}
    \label{fig:architecture}
\end{figure}

\begin{figure}[t!]
    \centering
    \includegraphics[width=1\linewidth]{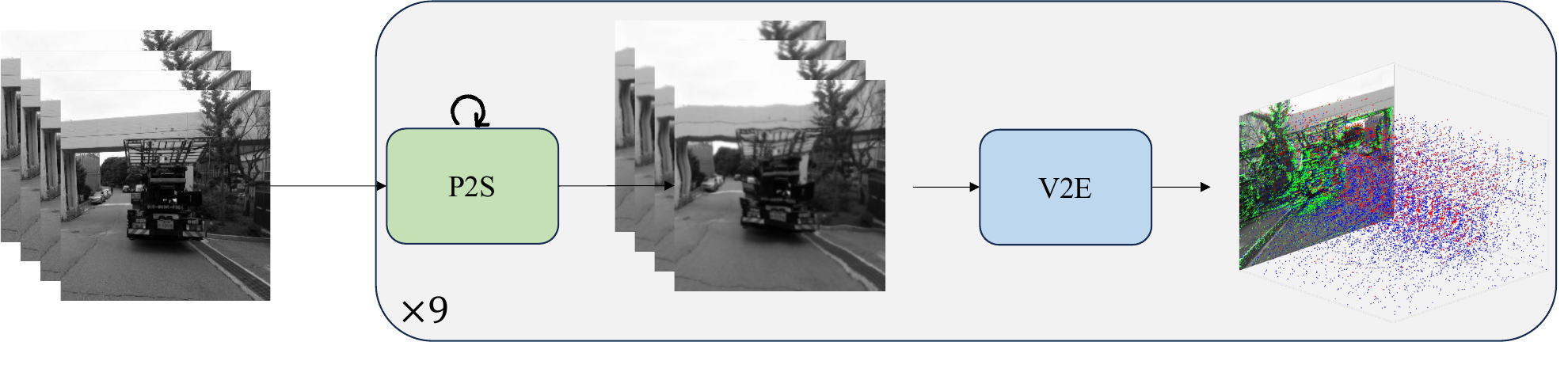}
    \caption{\textbf{Simulation pipeline.} Turbulence is added to clean frames using a turbulence simulator based on the phase-to-space (P2S) transform~\cite{turbulence_p2s}. V2E converts turbulent frames to events~\cite{v2e}. The process is repeated 9 times to simulate a $3 \times 3$ light field. The network is trained on simulated data and generalizes to tabletop experimental settings.}
    \label{fig:datagen}
\end{figure}

\subsection{Turbulence Mitigation Model}

While the $N$ different views provide critical cross-view constraints on the shared scene $S$, the system remains underdetermined without physical regularization on $h$.
Rather than designing explicit priors for this non-convex joint estimation problem, we adopt a data-driven approach: a neural network is trained to directly approximate the inverse mapping from the $N$ event streams to the clean video $V$.
By training on a large dataset of simulated turbulence~\cite{turbulence_p2s}, the network implicitly learns the spatial statistics of anisoplanatic turbulence and the manifold of natural scenes, leveraging cross-view consensus to resolve the bilinear ambiguity.

We implement this by adapting E2VID~\cite{e2vid} to the multi-view setting with minimal modification, so that the performance gain can be attributed to the light field geometry rather than architectural choices.
In its original single-view form, E2VID takes an event stream $\mathcal{E}$ as input and outputs a reconstructed video $\hat{V}$. E2VID represents input events as a spatio-temporal voxel grid with 5 temporal bins, whose values contain temporally weighted summed event polarity.

We extend E2VID to the multi-view setting by stacking each view's voxel representation along the temporal bin dimension as additional channels, forming a joint input
illustrated in \autoref{fig:architecture}.
E2VID processes the stacked input with a recurrent convolutional encoder-decoder, whose recurrent state accumulates temporal evidence across windows, producing the reconstructed frames $\hat{V}$.

\subsubsection{Training Details}
The network is trained to minimize the LPIPS loss~\cite{lpips} averaged over all reconstructed frames:
\begin{equation}
    \mathcal{L} = \frac{1}{K}\sum_{k=1}^{K}\operatorname{LPIPS}(\hat{V}_k, V_k).
\end{equation}
E2VID is traditionally trained with the calibrated perceptual loss LPIPS instead of mean squared error (MSE) because MSE losses often produce blurry images \cite{e2vid}. The network is trained from scratch for $200$ epochs with a batch size of $8$ using the Adam optimizer~\cite{Kingma2015Adam} with a constant learning rate of $10^{-4}$ on a single NVIDIA RTX A6000 GPU.

We follow the augmentation strategy of~\cite{e2vid,e2vid+}.
Geometric augmentation applies random $112 \times 112$ crops and flips with probability $0.5$.
Noise augmentation adds $\mathcal{N}(0, 0.4)$ to $\tilde{E}_k$ and injects hot-pixel noise $\mathcal{N}(0, 0.1)$ at up to $0.01\%$ of pixel locations.
Pause augmentation zeros input events with probability $0.05$ and maintains the paused state with probability $0.9$, training the recurrent state to preserve the output without new events.
Spectral normalization~\cite{yoshida2017spectralnorm} is applied to the ConvLSTM layers to prevent artifacts from accumulating as the recurrent hidden state propagates across temporal windows.

\section{Experiments}
\subsubsection{Simulated Dataset}
A synthetic light field event dataset is constructed by combining a physics-based turbulence simulator (P2S)~\cite{turbulence_p2s} with an event simulator (V2E)~\cite{v2e}, illustrated by \autoref{fig:datagen}.
Clean intensity frames are taken from the REDS dataset~\cite{Nah_2019_CVPR_Workshops_REDS}, whose 120\,fps videos provide high-frame-rate intensity sequences for event synthesis. The central region is cropped and resized to $256 \times 256$, serving as ground truth.
For each clip, $N=9$ independent turbulence realizations are generated by running the turbulence simulator with the same parameters but different random seeds for the P2S, producing nine sequences of degraded frames.
Each degraded sequence is then passed through the event simulator to produce the corresponding sub-aperture event stream $\mathcal{E}^{(i)}$. The contrast threshold $C$ is uniformly sampled from $[0.1, 0.7]$ and the refractory period is set to the default $5$\,ms.
The temporal window $\Delta T_k$ in the voxel grid is set to one frame interval.

To improve generalization across turbulence conditions, we sample turbulence parameters from the ATSyn dataset~\cite{datum}, which provides physically grounded turbulence parameters, and adapt P2S following the simulation procedure described in the same work.
First, the pixel pitch $p$ is computed from the focal length $f$, scene width, and propagation distance, and the geometric tilt is scaled by $f/p$.
Second, the blur kernel is resized to match the scale specified by the ATSyn parameters.
Finally, temporal correlation is introduced by relating the random seed $\boldsymbol{n}_t$ at time $t$ to its predecessor via $\boldsymbol{n}_t = \alpha\boldsymbol{n}_{t-1} + \sqrt{1-\alpha^2}\,\boldsymbol{\epsilon}_t$, where $\alpha$ is the temporal correlation coefficient and $\boldsymbol{\epsilon}_t \sim \mathcal{N}(\mathbf{0}, \mathbf{I})$.

The network is trained exclusively on this simulated dataset but is evaluated on both simulated and experimental data.
The dataset is split into 216 training, 27 validation, and 27 test clips, each elapsing 4 seconds.
Using the turbulence severity criteria in~\cite{datum}, our simulated dataset consists of 60\% strong, 30\% medium, and 10\% weak clips.
A $768 \times 768$ version of the dataset was generated to compare single-view reconstructions estimated from the same number of pixels as the $3 \times 3  \times 256  \times 256$ light field. Complete characterization of the light-field/resolution tradeoff is provided in the supplemental material.

\begin{figure}[t!]
    \centering
    \includegraphics[width=1\linewidth]{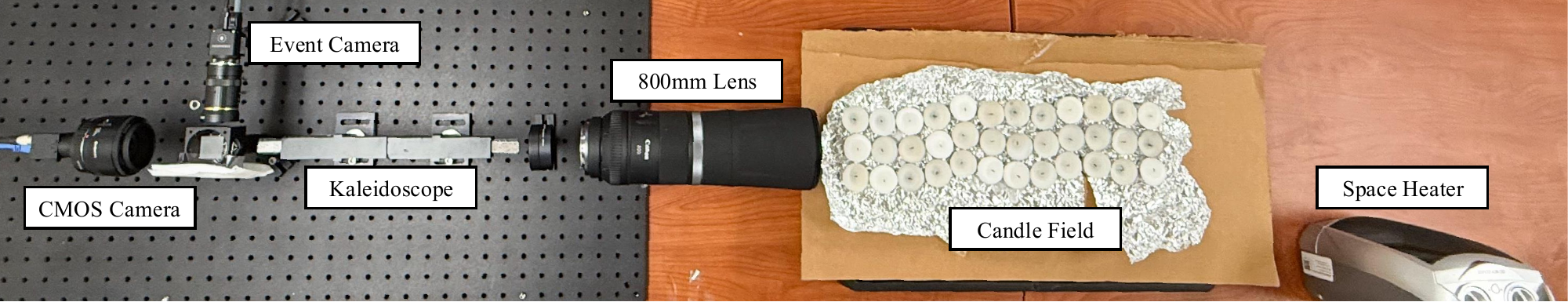}
    \caption{\textbf{Tabletop experimental setup.} Real turbulence is generated from temperature fluctuations caused by candles and a space heater. We placed a beam splitter behind the kaleidoscope to simultaneously capture events and reference CMOS frames.}
    \label{fig:experiment_setup}
\end{figure}

\subsubsection{Tabletop Experiments}
As illustrated in \autoref{fig:experiment_setup}, the light field event camera comprises two optical stages in series.
A Canon RF 800 mm f/11 main lens focuses the scene onto an intermediate image plane. To match the exit pupil of the main lens with the entrance pupil of the pickup system, a 100 mm pupil-matching relay lens is introduced. The pickup system forms an image from the intermediate plane onto the sensor plane. Through the reflections of a $305\,\mathrm{mm}$-long kaleidoscope with a $12.6\,\mathrm{mm}$ aperture, similar to the setup in \cite{qu2025eventfield}, a 3 × 3 array of sub-aperture views is formed.
A beam splitter behind the kaleidoscope directs light simultaneously to a Prophesee EVK4 event sensor for event capture and a Blackfly S BFS-U3-13Y3C CMOS camera for reference imagery.


Turbulence is generated in the laboratory by placing a $3 \times 12$ grid of tea candles starting $61.0\,\mathrm{mm}$ from the main lens and a $1500$ watt space heater $813\,\mathrm{mm}$ from the same lens, as illustrated in \autoref{fig:experiment_setup}.
All experiments are conducted with targets at a distance of $9.14\,\mathrm{m}$ from the camera. This is not far enough to assume targets are at optical infinity, and the kaleidoscope alignment is not perfect, so each sub-aperture view is aligned to the center with a homography estimated from a calibration target presented simultaneously to each view. 



\begin{figure}[t!]
    \centering
    \includegraphics[width=0.85\linewidth]{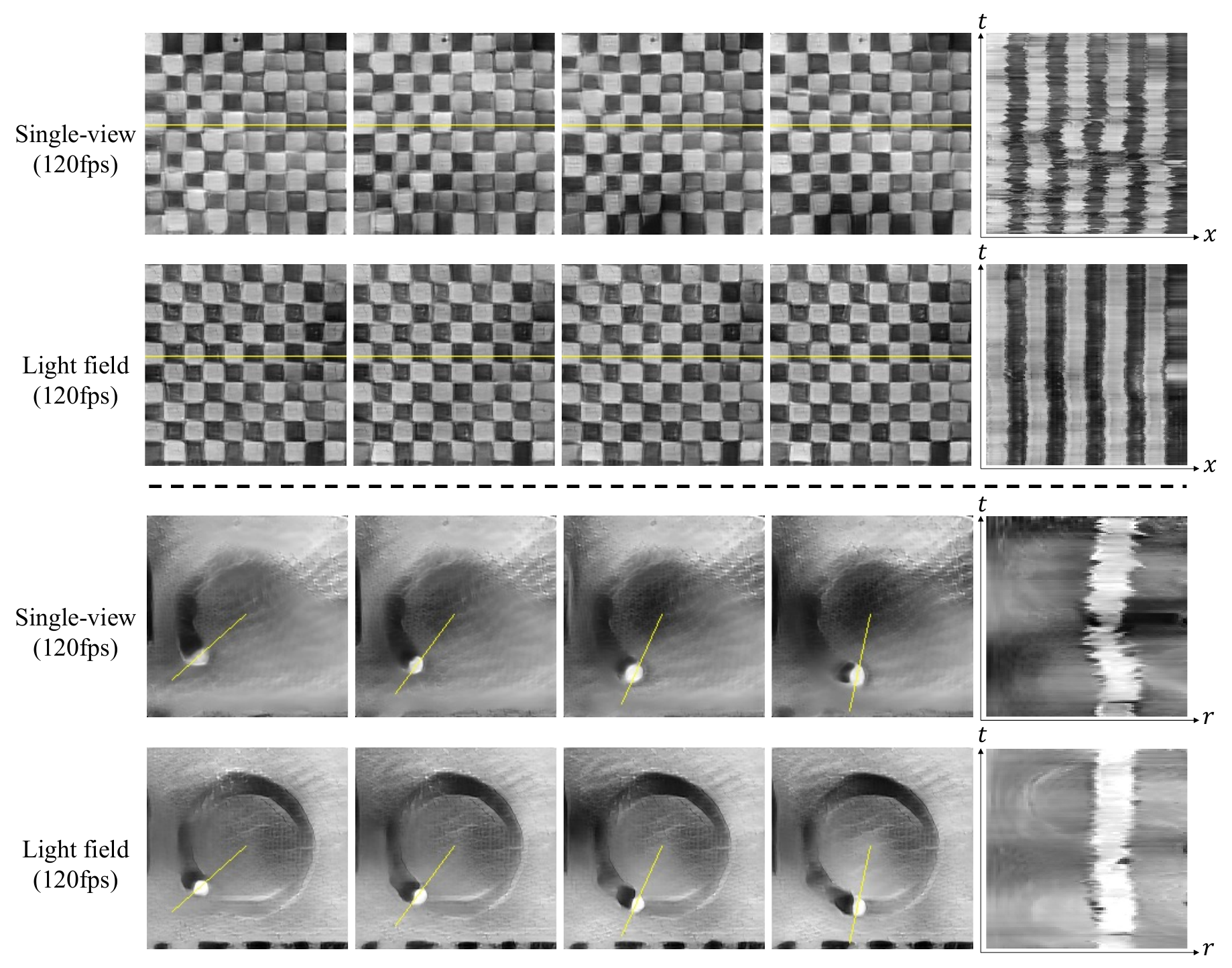}
    \caption{\textbf{Light fields improve temporal consistency on experimental data.} In the laboratory, we imaged a static checkerboard and a rotating dot, both displayed on an LCD monitor, through turbulence. The left four columns show reconstructed frames at $120$\,fps. The right column visualizes $x$-$t$ and $r$-$t$ plots showing the time evolution the slice marked in yellow. The light field results in straighter lines, indicating less wander.}
    \label{fig:temporal_consistency}
\end{figure}

\begin{table}[t]
\caption{\textbf{Simulated quantitative results.} On the REDS dataset, 
the light field improves frame quality in comparison to a single-view configuration.}
\label{tab:quantitative_results}
\centering
\begin{tabular}{l|c|ccccc} 
\toprule
{Method} & Resolution & PSNR $\uparrow$ & SSIM $\uparrow$ & LPIPS $\downarrow$ & ITF $\uparrow$ & $E_{\text{warp}}$ $\downarrow$ \\ 
\midrule
Single-view & $768 \times 768$ & $14.84$ & $0.501$ & $0.5185$ & $22.71$ & $0.0050$ \\
Single-view & $256 \times 256$ & $14.84$ & $0.461$ & $0.4607$ & $22.72$ & $0.0047$ \\
Light field & $3\times 3\times 256 \times 256$ & $\mathbf{15.40}$ & $\mathbf{0.505}$ & $\mathbf{0.4332}$ & $\mathbf{25.16}$ & $\mathbf{0.0022}$ \\
\bottomrule
\end{tabular}
\end{table}

\begin{table}[t]
\caption{\textbf{Tabletop experiment quantitative results.} 
Reconstructing REDS videos displayed through turbulence, light field outperforms single-view.}
\label{table:real_exp}
\centering
\footnotesize
\setlength{\tabcolsep}{2.5pt}
\renewcommand{\arraystretch}{1.05}
\begin{tabular}{l|cccccc}
\toprule
Method & PSNR $\uparrow$ & SSIM $\uparrow$ & LPIPS$ \downarrow$ & ITF $\uparrow$ & $E_{\text{warp}} \downarrow$ \\
\midrule
Single-view & $12.37$ & $0.351$ & $0.543$ & $20.87$ & $0.0085$ \\
Light field & $\mathbf{12.99}$ & $\mathbf{0.367}$ & $\mathbf{0.531}$  & $\mathbf{23.30}$ & $\mathbf{0.0051}$ \\
\bottomrule
\end{tabular}
\end{table}

\section{Results}
\label{sec:results}
\subsection{Event Light Field versus Single-View Reconstruction}

To demonstrate light field optics capture sufficient information to remove turbulence while preserving scene motion, we compare the light field configuration against a single-view baseline using the same reconstruction architecture trained on turbulence-degraded data from the central sub-aperture only.

\begin{figure}[t!]
    \centering
    \includegraphics[width=1\linewidth]{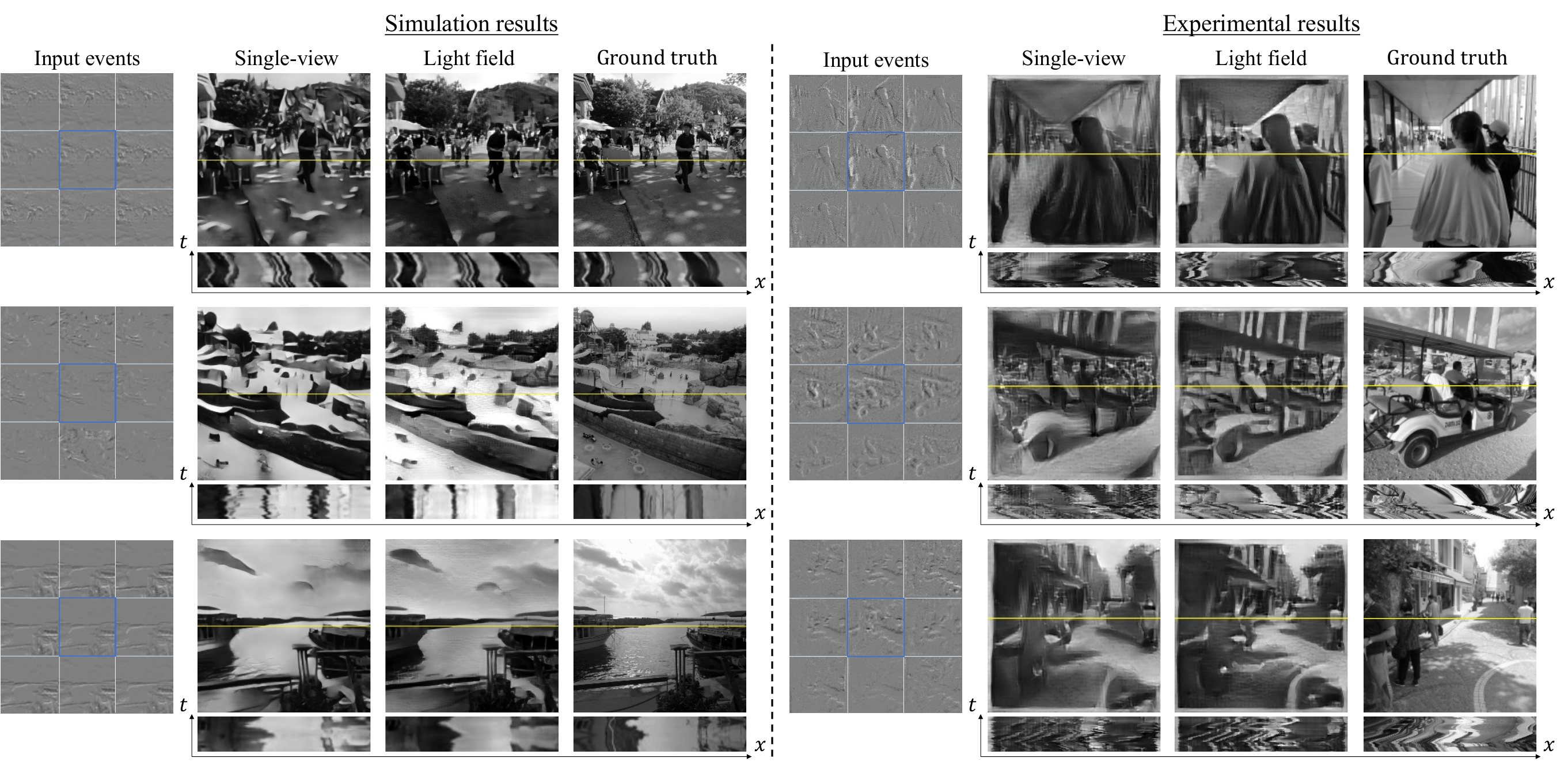}
    \caption{\textbf{Reconstruction with light field input outperforms single-view on simulated and experimental data.} Each group shows the input events (blue box indicates the central sub-aperture used for the single-view baseline), single-view reconstruction, the light field reconstruction, and the ground truth. Simulated single-view results are at $768 \times 768$ resolution, which has the same number of input pixels as the $3\times 3$ light field whose views are simulated at $256 \times 256$. Experimental single-view and light field results are at $176 \times 176$, with the target videos displayed on an LCD monitor. The light field model recovers sharper edges with fewer artifacts than the single-view baselines at either resolution. The $x$-$t$ plots below each group show the time evolution of the slice marked in yellow. Their relative smoothness shows that the light field reconstruction maintains a consistent appearance over time, while the single-view output exhibits frame-to-frame instability caused by turbulence. Videos in supplemental material more clearly illustrate turbulence reduction than images.}
    \label{fig:red_cv_lf}
\end{figure}

As illustrated in \autoref{fig:red_cv_lf}, the model trained on light field views reconstructs edges more faithfully than the single-view baseline. This is apparent along structural lines: the single-view approach bends them into curved artifacts, whereas the light field model preserves sharper, straighter edges. This improvement is particularly evident in real-world experiments, where the single-view baseline suffers from blurred edges and dimming artifacts. The $x$-$t$ plots further show that the light field reconstruction is more temporally consistent, with fewer frame-to-frame fluctuations caused by turbulence.

\autoref{tab:quantitative_results} quantitatively shows that the light field model outperforms the single-view baseline in both reconstruction quality and temporal stability. This holds even when the total pixel count is matched by using three times the spatial resolution in both the horizontal and vertical dimensions to match the aggregate pixel count of the $3 \times 3$ light field.
We evaluate temporal stability using Inter-frame Transformation Fidelity (ITF)~\cite{aguilar2014itf} and warping error ($E_{\text{warp}}$)~\cite{Lai-ECCV-2018-ewarp}; complete implementation details are provided in the supplemental material.

Additionally, two real-world experiments are presented in \autoref{fig:temporal_consistency} to illustrate how the light field approach reconstructs consistent videos at a traditional frame rate of 120 fps. Higher frame rate results are shown later.
First, a static checkerboard is reconstructed through a turbulent medium. In the $x$-$t$ plot, one can see that the single-view baseline exhibits clear temporal jitter, whereas the light field method maintains highly stable edges over time. Next, a spinning dot is displayed on a flicker-free, LCD monitor. While the monitor displays frames discretely at 120 Hz, the motion-blur inherent to LCD monitors smooths the scene's motion.
The $r$-$t$ plot is constructed by creating a slice that follows the dot's spin.
This plot demonstrates the light field configuration effectively reduces spatial wandering caused by turbulence.
Quantitative results for the tabletop experiments are reported in \autoref{table:real_exp}. The ground truth is a separate capture acquired with the same experimental setup but without turbulence.
Across metrics, the light field configuration consistently outperforms the single-view baselines.

\subsection{Event Light Fields Overcome Motion Blur}

\begin{figure}[t!]
    \centering
    \includegraphics[width=1\linewidth]{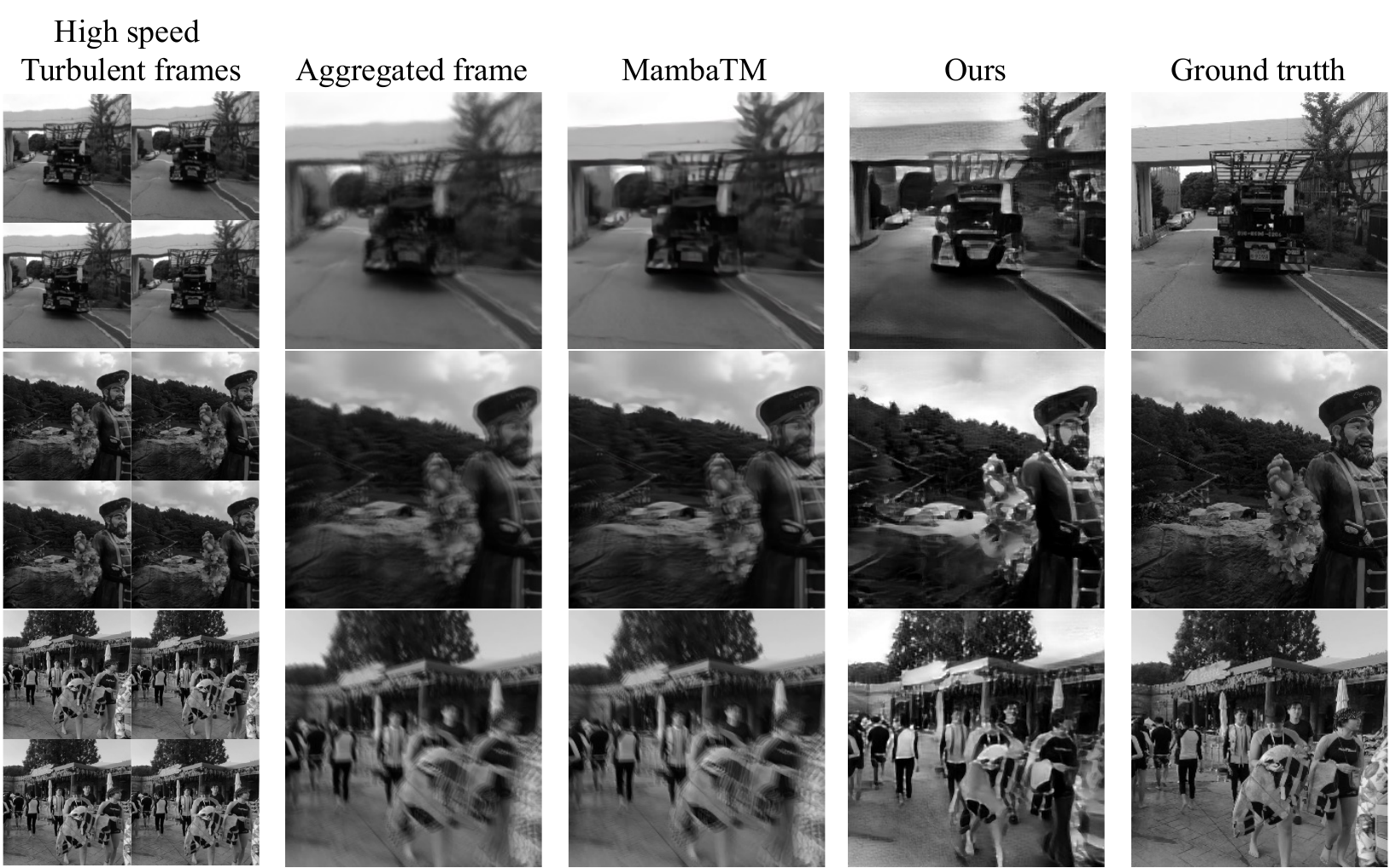}
    \caption{\textbf{Simulated high-speed scenes show that event light fields recover sharper images than frame-based methods.} Motion-blur cannot be compensated for by frame-based methods, which occurs when scene dynamics or turbulence are fast. Event-based light fields operate directly on the event stream and reconstruct the underlying scene without blur caused by integration. Motion-blur is simulated by aggregating 4 conventional frames into a single blurred frame.}
    \label{fig:mambatm_comp}
\end{figure}
We test the ability of event light fields to mitigate motion blur by simulating motion blur and comparing to MambaTM~\cite{mambatm}, the state-of-the-art frame-based turbulence mitigation method.
High speed imagery is simulated by averaging each group of four consecutive frames.
As illustrated in~\autoref{fig:mambatm_comp}, MambaTM reconstructions still contain blur and distortions.
Our method, in contrast, ingests event streams, which are free from the camera integration window, simulated from the same high-speed frames, and reconstructs video directly. The most relevant prior methods for single-view turbulence mitigation using events, EvTurb~\cite{evturb} and EGTM~\cite{egtm} have not publicly released code, precluding direct comparison. These methods use both frames and events.

\begin{figure}[ht!]
    
        \centering
        \includegraphics[width=0.71\textwidth]{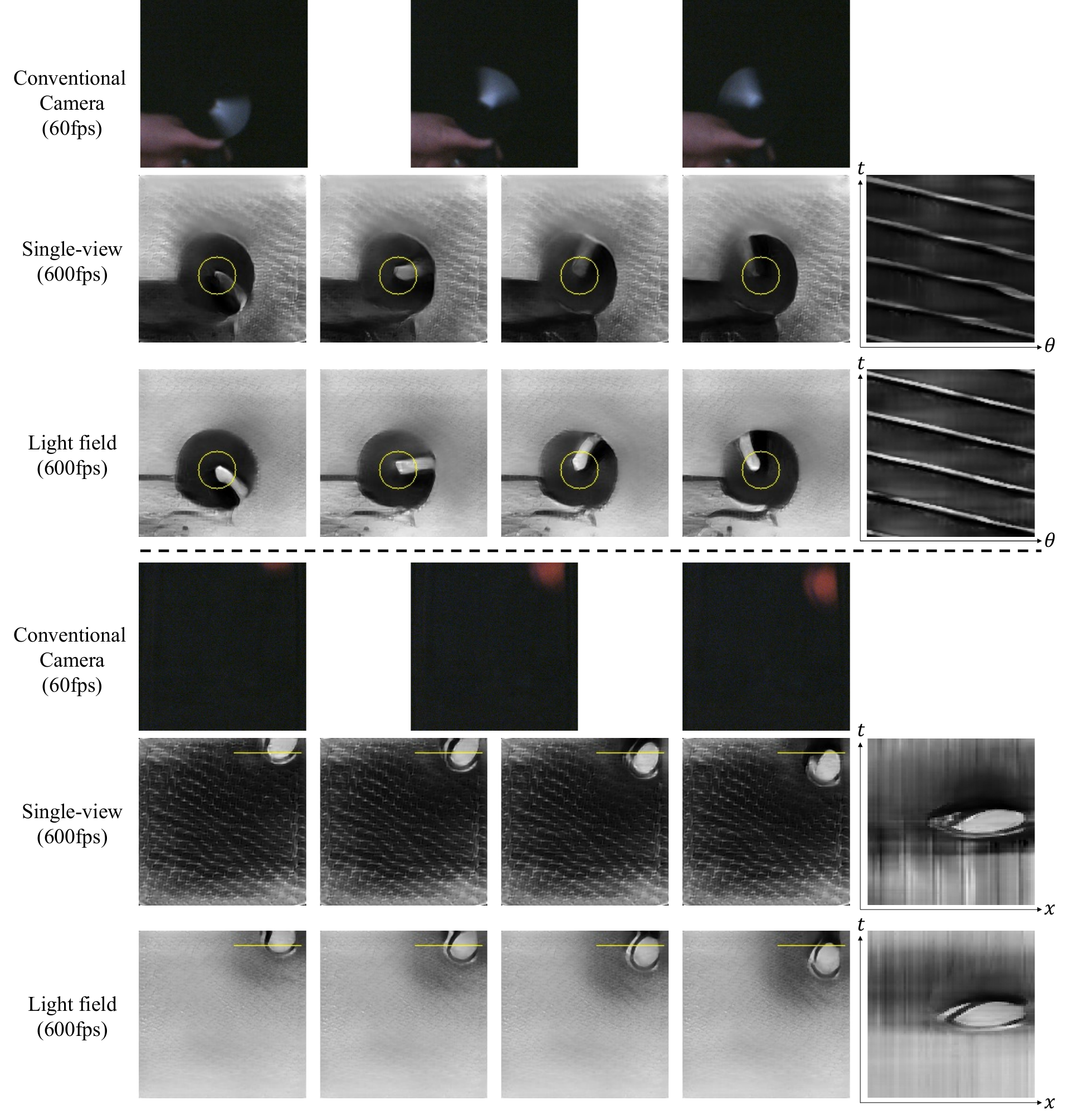}
        \label{fig:high_speed_ball}
    
    \caption{\textbf{Event-based light field design corrects turbulence in experimental high-speed regimes.} In lab, we imaged a spinning reflective stripe (top) and a bouncing ball (bottom). A $600$\,fps video is reconstructed, with samples shown four frames apart. Our method successfully reconstructs frame detail while correcting for wobble, as shown in the last column. The $\theta$-$t$ and $x$-$t$ plots show a $200$\,ms interval. Videos in supplemental material more clearly illustrate turbulence reduction than images.}
    \label{fig:high_speed_x_t}
\end{figure}

\begin{figure}[ht!]

    \centering
    \includegraphics[width=0.8\linewidth]{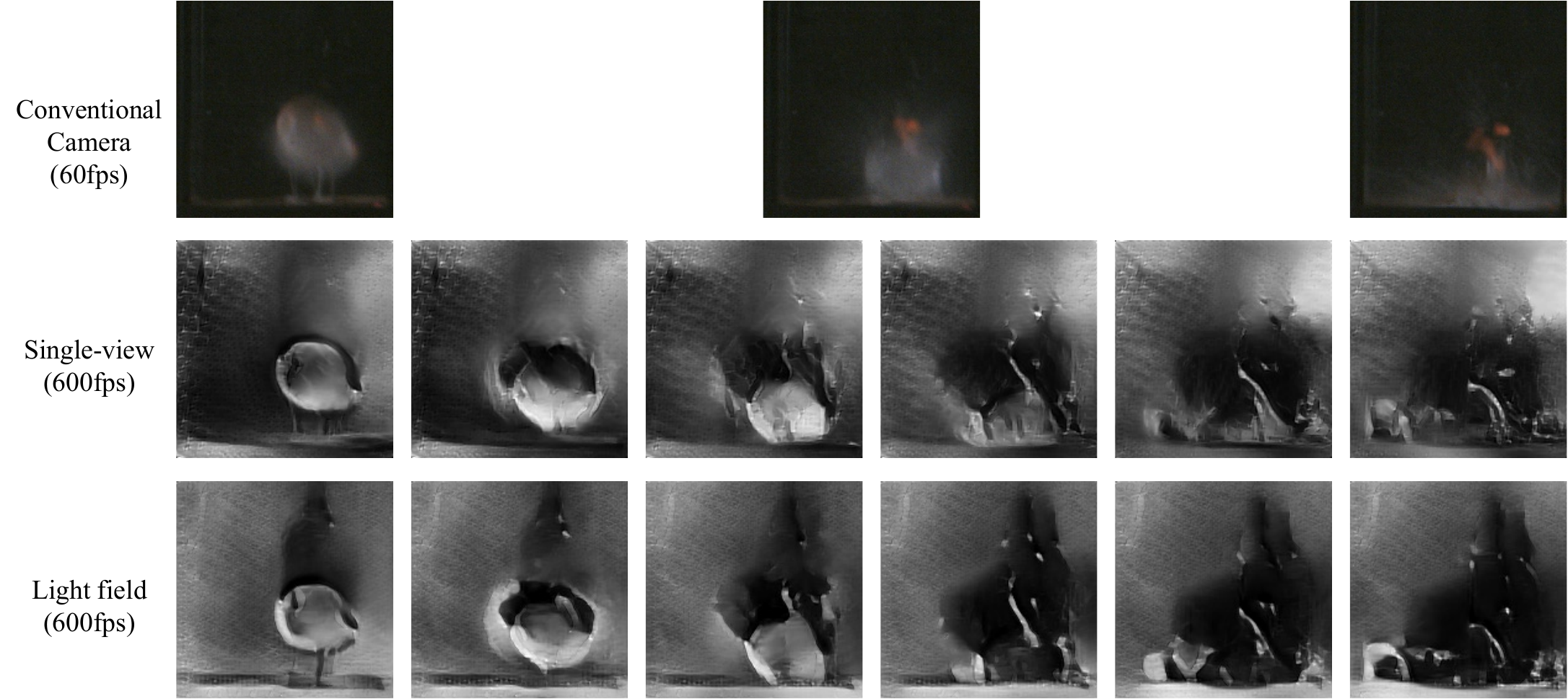}
    \caption{\textbf{High-speed video reconstruction in experimental turbulent scenes with complex dynamics.} In lab, our method recovers a water balloon bursting on top of pins at $600$\,fps. The displayed images are spaced 5 frames apart. Please see the videos included in the supplemental material for a clearer comparison.}
    \label{fig:high_speed_real}
\end{figure}

\subsection{Recovering High-Speed Video Through Turbulence}


We performed multiple in-lab, tabletop experiments to demonstrate that event-based light field imaging enables high-speed long-range imaging through turbulence. Unless noted otherwise, we reconstruct video at $600$\,fps.
As shown in~\autoref{fig:high_speed_x_t}, we captured a spinning stripe and a bouncing ball in lab.
Since the spinning stripe rotates uniformly, the $\theta$-$t$ plot should show straight sloped lines. The single-view configuration produces a wavy trajectory, revealing residual turbulence corruption; the light field reconstruction recovers a  straighter slope, consistent with the expectation.
For the bouncing ball, the single-view $x$-$t$ plot shows geometric distortion and loss of surface texture, while the light field reconstruction preserves the ball's texture along the full cross-sectional trajectory.

\begin{figure}
    \centering
    \includegraphics[width=0.95\linewidth]{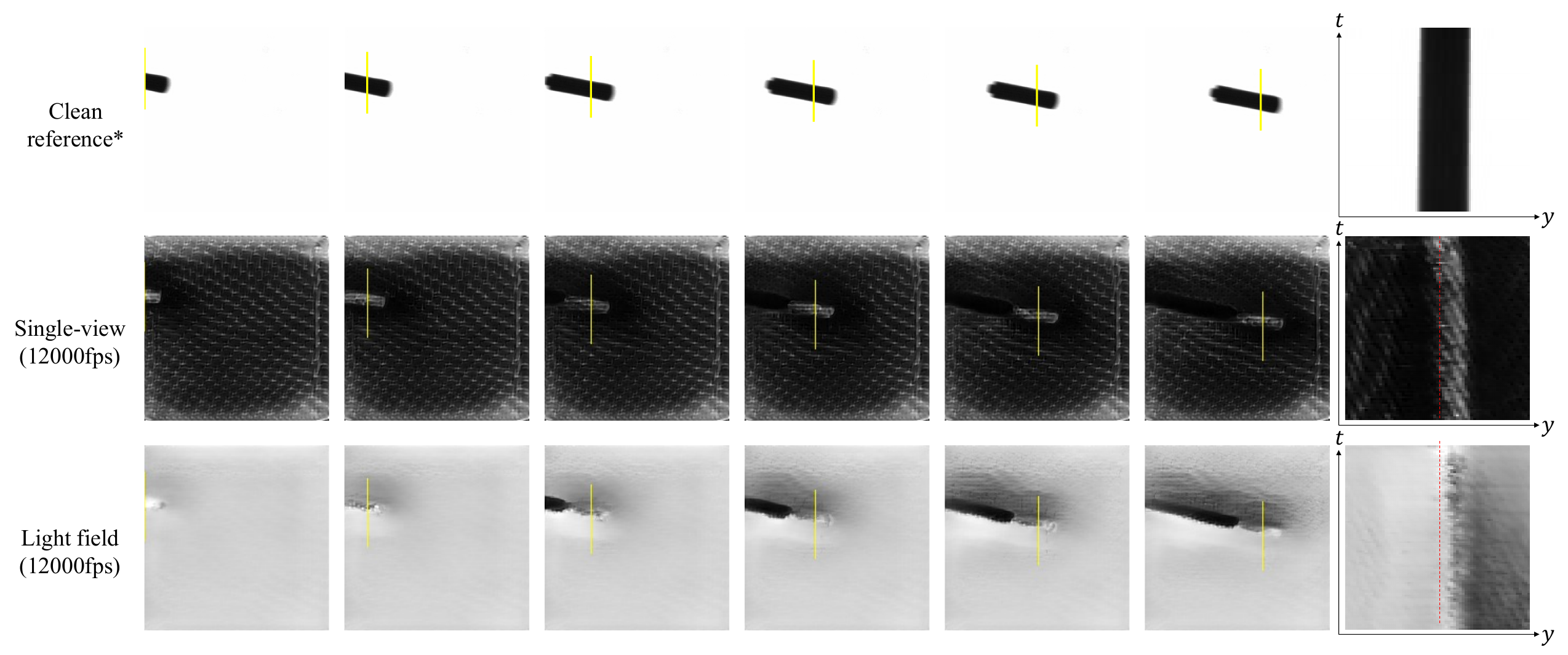}
    \caption{\textbf{Experimental reconstruction of a Nerf dart at 12,000 fps.} The first row shows a turbulence-free HSC capture, where a Nerf dart follows a straight trajectory. The second two rows show reconstructions of a dart taking a similar, but not identical, trajectory imaged through turbulence with the event camera. Compared with the single-view baseline, the light-field model more accurately reconstructs the straight path of a passing dart traveling at 16,000 pixels per second. The $y$-$t$ plot shows 8.3 ms of data, with images spaced 16 frames apart.
}
    \label{fig:nerf}
\end{figure}

Additionally, as shown in~\autoref{fig:high_speed_real}, we captured a water balloon bursting. The light field configuration captures the complex water dynamics without confusing them with turbulence. Finally, as shown in~\autoref{fig:nerf}, we captured a Nerf dart traveling at 16,000 pixels per second. We successfully estimate the video at 12,000\,fps, with the light field reconstructing a noticeably clearer and more accurate straight line trajectory when compared to a reference captured with a high-speed camera (Photron UX100) recording at 4,000 fps with additional back-lighting.

\

\subsection{Comparison with High-Speed Camera}
\begin{figure}[t!]
    \centering
    \includegraphics[width=0.95\linewidth]{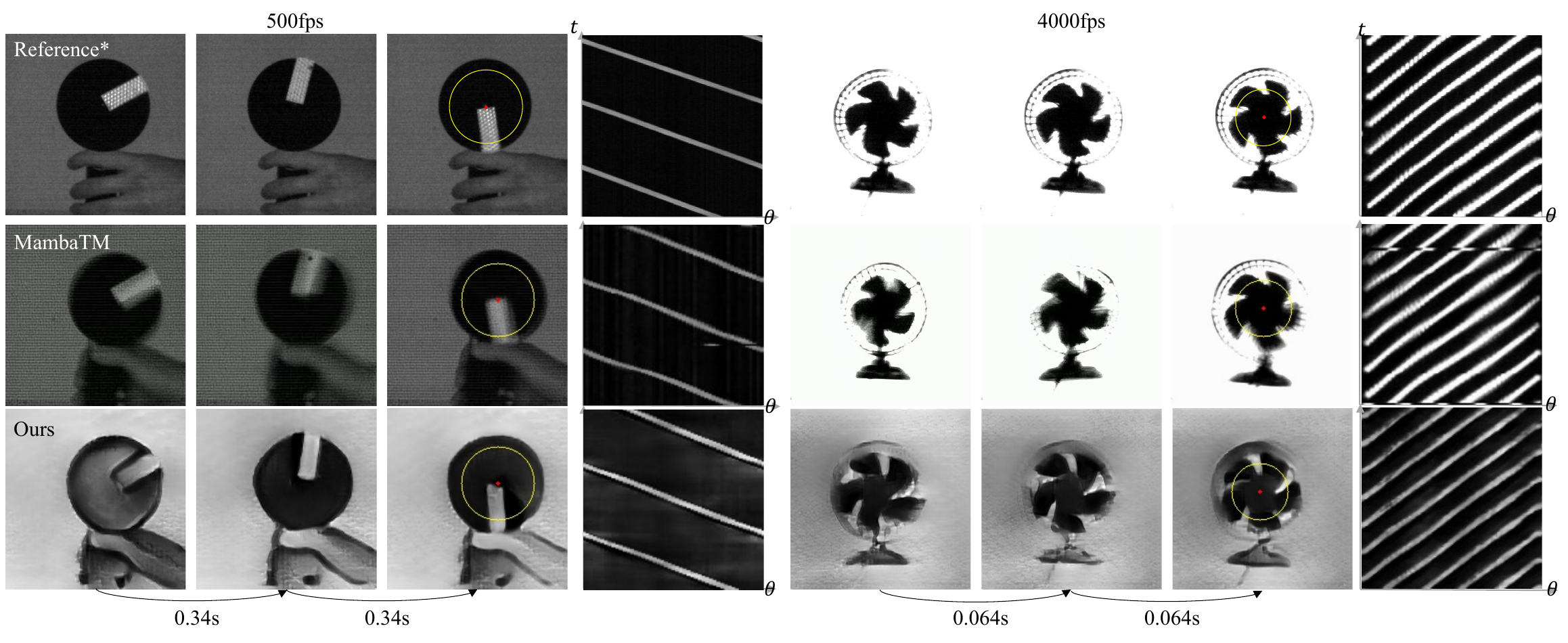}
    \caption{\textbf{Comparison with high-speed camera.} With a \$100k, high-speed camera MambaTM retains residual distortion, while ours, at $\sim$20$\times$ lower cost, stays sharp. Note that the reference was captured separately from the turbulent HSC and event data, and therefore temporally post-aligned.}
    \label{fig:hsc}
\end{figure}

To our knowledge, ours is the first event-only turbulence mitigation method for dynamic scenes, leaving no directly comparable high-speed baseline. The closest event-based methods, EvTurb~\cite{evturb} and EGTM~\cite{egtm}, require both events and frames, target static scenes, and have no released code. We therefore compare against MambaTM paired with a Photron UX100 high-speed camera (HSC) operating at 500--4000\,fps. This camera is an order of magnitude more expensive (\$100k vs. \$5k), far heavier (1.5\,kg vs. 40\,g camera body), has a limited capture duration (2\,s maximum vs. unlimited), and requires substantially more light (targets had to be lit with 2--45$\times$ more light than previous experiments). As illustrated in~\cref{fig:hsc},  this powerful baseline still exhibits residual blur and distortion compared to our method. MambaTM's batching also causes glitches, visible in the $\theta$-$t$ plots.

\section{Limitations}

Our system inherits the limited spatial resolution of event cameras ($\sim$1 MP maximum) and further trades spatial resolution for angular diversity. A multi-camera light field array avoids this tradeoff, but would increase cost and require synchronization across sensors. The current design assumes that the scene is approximately at optical infinity, with limited parallax between sub-aperture views. This is reasonable for long-range atmospheric imaging settings, but does not hold for close-range applications such as biomedical imaging through optical aberrations. 
Generalization beyond laboratory scenes remains an open challenge, particularly for real-world high-dynamic-range scenes where event simulation can be less accurate.
The current reconstruction algorithm has a latency of 3.4\,ms at $176 \times 176$ resolution. This supports real-time operation at the evaluated resolution, but higher-resolution deployment requires further optimization.

\section{Conclusion}

We introduced and demonstrated an approach to recovering high-speed video through atmospheric turbulence using a light field event camera, 
which does not rely upon and is not constrained by a conventional frame-based sensor. 
The event camera offers high temporal resolution to resolve fast dynamics. However, on its own, it cannot discern scene dynamics from turbulence.
We showed that a light field design resolves this issue. Each sub-aperture sees the same scene through a different  path in the atmosphere. Then, the scene signal is consistent across views while the turbulence signature varies. Experiments with neural network based event to video reconstruction confirm that event fields mitigate turbulence better than a single-view, event fields overcome the motion-blur inherent to frame-based turbulence mitigation, simulated event fields can be used to train models that work on real-data, and event fields can reconstruct complex scene dynamics without confusing them with turbulence.




\section*{Acknowledgements}
We gratefully acknowledge support from the Maryland Robotics Center. L.B. was supported by the TRX Systems Postdoctoral Fellowship. Z.Q. and A.P. were supported in part by the National Science Foundation under Grant No.~2403122. S.S.~was supported by an NDSEG Fellowship through the Air Force Office of Scientific Research under award number FA955025CB010 in the amount of \$153,984. Y.H.,~L.B.,~and C.A.M.~were supported in part by ARO award no. W911NF2420113, ONR award no.~N000142312752, and NSF CAREER award no.~2339616.
This work is also supported in part by the Defense Advanced Research Projects Agency (DARPA) under the CIDAR program.

%
%
\bibliographystyle{splncs04}
\bibliography{main}

@String(CVPR  = {IEEE Conf. Comput. Vis. Pattern Recog.})

@String(ICCV  = {Int. Conf. Comput. Vis.})

@String(ICLR  = {Int. Conf. Learn. Represent.})

@String(CVPRW = {IEEE Conf. Comput. Vis. Pattern Recog. Worksh.})

@String(AAAI  = {AAAI})

@String(CVPR  = {CVPR})

@String(ICCV  = {ICCV})

@String(ICLR  = {ICLR})

@String(CVPRW = {CVPRW})

@article{hampson2021adaptive,
  title={Adaptive optics for high-resolution imaging},
  author={Hampson, Karen M and Turcotte, Rapha{\"e}l and Miller, Donald T and Kurokawa, Kazuhiro and Males, Jared R and Ji, Na and Booth, Martin J},
  journal={Nature Reviews Methods Primers},
  volume={1},
  number={1},
  pages={68},
  year={2021},
  publisher={Nature Publishing Group UK London}
}

@inproceedings{datum,
  title={Spatio-Temporal Turbulence Mitigation: A Translational Perspective},
  author={Zhang, Xingguang and Chimitt, Nicholas and Chi, Yiheng and Mao, Zhiyuan and Chan, Stanley H},
  booktitle={2024 IEEE/CVF Conference on Computer Vision and Pattern Recognition (CVPR)},
  pages={2889--2899},
  year={2024},
  organization={IEEE Computer Society}
}

@inproceedings{mambatm,
  title={Learning Phase Distortion with Selective State Space Models for Video Turbulence Mitigation},
  author={Zhang, Xingguang and Chimitt, Nicholas and Wang, Xijun and Yuan, Yu and Chan, Stanley H},
  booktitle={Proceedings of the Computer Vision and Pattern Recognition Conference},
  pages={2127--2138},
  year={2025}
}

@article{tmt,
  title={Imaging through the atmosphere using turbulence mitigation transformer},
  author={Zhang, Xingguang and Mao, Zhiyuan and Chimitt, Nicholas and Chan, Stanley H},
  journal={IEEE Transactions on Computational Imaging},
  volume={10},
  pages={115--128},
  year={2024},
  publisher={IEEE}
}

@article{cai2024temporally,
  title={Temporally consistent atmospheric turbulence mitigation with neural representations},
  author={Cai, Haoming and Chen, Jingxi and Feng, Brandon Y and Jiang, Weiyun and Xie, Mingyang and Zhang, Kevin and Fermuller, Cornelia and Aloimonos, Yiannis and Veeraraghavan, Ashok and Metzler, Christopher A},
  journal={Advances in Neural Information Processing Systems},
  volume={37},
  pages={44554--44574},
  year={2024}
}

@inproceedings{nert,
  title={Ne{RT}: Implicit neural representations for unsupervised atmospheric turbulence mitigation},
  author={Jiang, Weiyun and Boominathan, Vivek and Veeraraghavan, Ashok},
  booktitle={Proceedings of the IEEE/CVF Conference on Computer Vision and Pattern Recognition},
  pages={4236--4243},
  year={2023}
}

@article{evturb,
  title={Ev{T}urb: Event camera guided turbulence removal},
  author={Liu, Yixing and Teng, Minggui and Xia, Yifei and Duan, Peiqi and Shi, Boxin},
  journal={IEEE Transactions on Computational Imaging},
  year={2026},
  publisher={IEEE}
}

@article{egtm,
  title={{EGTM}: Event-guided Efficient Turbulence Mitigation},
  author={Li, Huanan and Fan, Rui and Guan, Juntao and Hao, Weidong and Rui, Lai and Wu, Tong and Wang, Yikai and Gu, Lin},
  journal={arXiv preprint arXiv:2509.03808},
  year={2025}
}

@inproceedings{boehrer2019eventatm,
  title={Using event cameras for imaging through atmospheric turbulence},
  author={Boehrer, Nicolas and Nieuwenhuizen, Robert and Dijk, Judith},
  booktitle={COAT-2019-workshop (Communications and Observations through Atmospheric Turbulence: characterization and mitigation)},
  year={2019}
}

@article{e2vid,
  title={High Speed and High Dynamic Range Video with an Event Camera.},
  author={Rebecq, H and Ranftl, R and Koltun, V and Scaramuzza, D},
  journal={IEEE Transactions on Pattern Analysis and Machine Intelligence},
  volume={43},
  number={6},
  pages={1964--1980},
  year={2019}
}

@inproceedings{firenet,
  title={Fast image reconstruction with an event camera},
  author={Scheerlinck, Cedric and Rebecq, Henri and Gehrig, Daniel and Barnes, Nick and Mahony, Robert and Scaramuzza, Davide},
  booktitle={Proceedings of the IEEE/CVF Winter Conference on Applications of Computer Vision},
  pages={156--163},
  year={2020}
}

@inproceedings{e2vid+,
  title={Reducing the sim-to-real gap for event cameras},
  author={Stoffregen, Timo and Scheerlinck, Cedric and Scaramuzza, Davide and Drummond, Tom and Barnes, Nick and Kleeman, Lindsay and Mahony, Robert},
  booktitle={European Conference on Computer Vision},
  pages={534--549},
  year={2020},
  organization={Springer}
}

@article{ercan2024hypere2vid,
  title={Hyper{E2VID}: Improving event-based video reconstruction via hypernetworks},
  author={Ercan, Burak and Eker, Onur and Saglam, Canberk and Erdem, Aykut and Erdem, Erkut},
  journal={IEEE Transactions on Image Processing},
  volume={33},
  pages={1826--1837},
  year={2024},
  publisher={IEEE}
}

@inproceedings{etnet,
  title={Event-based video reconstruction using transformer},
  author={Weng, Wenming and Zhang, Yueyi and Xiong, Zhiwei},
  booktitle={Proceedings of the IEEE/CVF International Conference on Computer Vision},
  pages={2563--2572},
  year={2021}
}

@article{ge2025eventmamba,
  title={Event{M}amba: Enhancing Spatio-Temporal Locality with State Space Models for Event-Based Video Reconstruction},
  author={Ge, Chengjie and Fu, Xueyang and He, Peng and Wang, Kunyu and Cao, Chengzhi and Zha, Zheng-Jun},
  journal={Proceedings of the AAAI Conference on Artificial Intelligence},
  volume={39},
  number={3},
  pages={3104--3112},
  year={2025}
}

@inproceedings{qu2025eventfield,
  title={Event fields: Capturing light fields at high speed, resolution, and dynamic range},
  author={Qu, Ziyuan and Zou, Zihao and Boominathan, Vivek and Chakravarthula, Praneeth and Pediredla, Adithya},
  booktitle={Proceedings of the Computer Vision and Pattern Recognition Conference},
  pages={26910--26920},
  year={2025}
}

@article{he2024microsaccade,
  title={Microsaccade-inspired event camera for robotics},
  author={He, Botao and Wang, Ze and Zhou, Yuan and Chen, Jingxi and Singh, Chahat Deep and Li, Haojia and Gao, Yuman and Shen, Shaojie and Wang, Kaiwei and Cao, Yanjun and others},
  journal={Science Robotics},
  volume={9},
  number={90},
  pages={eadj8124},
  year={2024},
  publisher={American Association for the Advancement of Science}
}

@article{boehrer2021turbulence,
  title={Turbulence mitigation in imagery including moving objects from a static event camera},
  author={Boehrer, Nicolas and Nieuwenhuizen, Robert PJ and Dijk, Judith},
  journal={Optical Engineering},
  volume={60},
  number={5},
  pages={053101--053101},
  year={2021},
  publisher={Society of Photo-Optical Instrumentation Engineers}
}

@article{feng2023turbugan,
  title={Turbugan: An adversarial learning approach to spatially-varying multiframe blind deconvolution with applications to imaging through turbulence},
  author={Feng, Brandon Y and Xie, Mingyang and Metzler, Christopher A},
  journal={IEEE Journal on Selected Areas in Information Theory},
  volume={3},
  number={3},
  pages={543--556},
  year={2023},
  publisher={IEEE}
}

@inproceedings{yang2002camarray,
author = {Yang, Jason C. and Everett, Matthew and Buehler, Chris and McMillan, Leonard},
title = {A real-time distributed light field camera},
year = {2002},
isbn = {1581135343},
publisher = {Eurographics Association},
address = {Goslar, DEU},
booktitle = {Proceedings of the 13th Eurographics Workshop on Rendering},
pages = {77–86},
numpages = {10},
location = {Pisa, Italy},
series = {EGRW '02}
}

@article{manakov2013reconfigurable,
  title={A reconfigurable camera add-on for high dynamic range, multispectral, polarization, and light-field imaging},
  author={Manakov, Alkhazur and Restrepo, John and Klehm, Oliver and Hegedus, Ramon and Eisemann, Elmar and Seidel, Hans-Peter and Ihrke, Ivo},
  journal={ACM Transactions on Graphics},
  volume={32},
  number={4},
  pages={47--1},
  year={2013}
}

@article{fried1978probabilitylucky,
  title={Probability of getting a lucky short-exposure image through turbulence},
  author={Fried, David L},
  journal={Journal of the Optical Society of America},
  volume={68},
  number={12},
  pages={1651--1658},
  year={1978},
  publisher={Optical Society of America}
}

@InProceedings{turbulence_p2s,
author = {Zhiyuan Mao and Nicholas Chimitt and Stanley H. Chan},
title = {Accelerating Atmospheric Turbulence Simulation via Learned Phase-to-Space Transform},
booktitle = {Proceedings of the IEEE/CVF International Conference on Computer Vision (ICCV)},
month = {October},
year = {2021}
}

@INPROCEEDINGS{v2e,

  title     = "v2e: From Video Frames to Realistic {DVS} Events",

  booktitle = "2021 {IEEE/CVF} Conference on Computer Vision and Pattern

               Recognition Workshops ({CVPRW})",

  author    = "Hu, Y and Liu, S C and Delbruck, T",

  publisher = "IEEE",

  year      =  2021

}

@inproceedings{lpips,
  title={The Unreasonable Effectiveness of Deep Features as a Perceptual Metric},
  author={Zhang, Richard and Isola, Phillip and Efros, Alexei A and Shechtman, Eli and Wang, Oliver},
  booktitle={Proceedings of the IEEE Conference on Computer Vision and Pattern Recognition (CVPR)},
  year={2018}
}

@article{wang2024neuromorphic,
  title={Neuromorphic shack-hartmann wave normal sensing},
  author={Wang, Chutian and Zhu, Shuo and Zhang, Pei and Huang, Jianqing and Wang, Kaiqiang and Lam, Edmund Y},
  journal={arXiv preprint arXiv:2404.15619},
  year={2024}
}

@article{pellizzari2026speckle,
  title={Speckle-robust digital adaptive optics for coherent reflective imaging},
  author={Pellizzari, Casey and Strong, David and Hardy, Tyler and Metzler, Chris and Spencer, Mark},
  year={2026},
  publisher={Optica Open}
}

@article{watnik2018wavefront,
  title={Wavefront sensing in deep turbulence},
  author={Watnik, Abbie T and Gardner, Dennis F},
  journal={Optics and Photonics News},
  volume={29},
  number={10},
  pages={38--45},
  year={2018},
  publisher={OSA}
}

@article{chimitt2025wavefront,
  title={Wavefront Estimation From a Single Measurement: Uniqueness and Algorithms},
  author={Chimitt, Nicholas and Almuallem, Ali and Guo, Qi and Chan, Stanley H},
  journal={IEEE Transactions on Computational Imaging},
  volume={11},
  pages={1600--1613},
  year={2025},
  publisher={IEEE}
}

@inproceedings{xie2024wavemo,
  title={Wave{M}o: learning wavefront modulations to see through scattering},
  author={Xie, Mingyang and Guo, Haiyun and Feng, Brandon Y and Jin, Lingbo and Veeraraghavan, Ashok and Metzler, Christopher A},
  booktitle={Proceedings of the IEEE/CVF Conference on Computer Vision and Pattern Recognition},
  pages={25276--25285},
  year={2024}
}

@article{yeminy2021guidestar,
  title={Guidestar-free image-guided wavefront shaping},
  author={Yeminy, Tomer and Katz, Ori},
  journal={Science advances},
  volume={7},
  number={21},
  pages={eabf5364},
  year={2021},
  publisher={American Association for the Advancement of Science}
}

@article{feng2023neuws,
  title={Neu{WS}: Neural wavefront shaping for guidestar-free imaging through static and dynamic scattering media},
  author={Feng, Brandon Y and Guo, Haiyun and Xie, Mingyang and Boominathan, Vivek and Sharma, Manoj K and Veeraraghavan, Ashok and Metzler, Christopher A},
  journal={Science Advances},
  volume={9},
  number={26},
  pages={eadg4671},
  year={2023},
  publisher={American Association for the Advancement of Science}
}

@article{jiang2026guidestar,
  title={Guidestar-free adaptive optics with asymmetric apertures},
  author={Jiang, Weiyun and Guo, Haiyun and Metzler, Christopher A and Veeraraghavan, Ashok},
  journal={ACM Transactions on Graphics},
  year={2026},
  publisher={ACM New York, NY}
}

@article{kong2020shack,
  title={Shack-Hartmann wavefront sensing using spatial-temporal data from an event-based image sensor},
  author={Kong, Fanpeng and Lambert, Andrew and Joubert, Damien and Cohen, Gregory},
  journal={Optics Express},
  volume={28},
  number={24},
  pages={36159--36175},
  year={2020},
  publisher={Optical Society of America}
}

@article{grose2024convolutional,
  title={Convolutional neural network for improved event-based Shack-Hartmann wavefront reconstruction},
  author={Grose, Mitchell and Schmidt, Jason D and Hirakawa, Keigo},
  journal={Applied Optics},
  volume={63},
  number={16},
  pages={E35--E47},
  year={2024},
  publisher={Optica Publishing Group}
}

@inproceedings{ziemann2024learning,
  title={A learning-based approach to event-based shack-hartmann wavefront sensing},
  author={Ziemann, Matthew R and Rathbun, Isabelle A and Metzler, Christopher A},
  booktitle={Unconventional Imaging, Sensing, and Adaptive Optics 2024},
  volume={13149},
  pages={199--207},
  year={2024},
  organization={SPIE}
}

@InProceedings{Nah_2019_CVPR_Workshops_REDS,
  author = {Nah, Seungjun and Baik, Sungyong and Hong, Seokil and Moon, Gyeongsik and Son, Sanghyun and Timofte, Radu and Lee, Kyoung Mu},
  title = {{NTIRE} 2019 Challenge on Video Deblurring and Super-Resolution: Dataset and Study},
  booktitle = {CVPR Workshops},
  month = {June},
  year = {2019}
}

@book{goodman2005introduction,
  title={Introduction to Fourier Optics},
  author={Goodman, J.W.},
  isbn={9780070242548},
  lccn={68017184},
  series={Electrical Engineering Series},
  year={1996},
  publisher={McGraw-Hill}
}

@article{Mao2020tmstanley,
  title={Image Reconstruction of Static and Dynamic Scenes Through Anisoplanatic Turbulence},
  author={Zhiyuan Mao and Nicholas Chimitt and Stanley H. Chan},
  journal={IEEE Transactions on Computational Imaging},
  year={2020},
  volume={6},
  pages={1415-1428}
}

@InProceedings{Li_2021_unsupervised,
    author    = {Li, Nianyi and Thapa, Simron and Whyte, Cameron and Reed, Albert W. and Jayasuriya, Suren and Ye, Jinwei},
    title     = {Unsupervised Non-Rigid Image Distortion Removal via Grid Deformation},
    booktitle = {Proceedings of the IEEE/CVF International Conference on Computer Vision (ICCV)},
    month     = {October},
    year      = {2021},
    pages     = {2522-2532}
}

@InProceedings{Kingma2015Adam,
  author    = {Kingma, Diederik and Ba, Jimmy},
  booktitle = {International Conference on Learning Representations (ICLR)},
  title     = {Adam: A Method for Stochastic Optimization},
  year      = {2015},
  address   = {San Diego, CA, USA},
  optmonth  = {12},
}

@article{chan2023turbulence_textbook,
  title={Computational imaging through atmospheric turbulence},
  author={Chan, Stanley H and Chimitt, Nicholas},
  journal={Foundations and Trends in Computer Graphics and Vision},
  volume={15},
  number={4},
  pages={253--508},
  year={2023},
  publisher={Emerald Publishing Limited}
}

@article{Wu2022integratedsensor,
	author = {Wu, Jiamin and Guo, Yuduo and Deng, Chao and Zhang, Anke and Qiao, Hui and Lu, Zhi and Xie, Jiachen and Fang, Lu and Dai, Qionghai},
	journal = {Nature},
	number = {7938},
	pages = {62--71},
	title = {An integrated imaging sensor for aberration-corrected 3D photography},
	volume = {612},
	year = {2022}}

@phdthesis{ng2005lightfieldcamera,
  title={Light field photography with a hand-held plenoptic camera},
  author={Ng, Ren and Levoy, Marc and Br{\'e}dif, Mathieu and Duval, Gene and Horowitz, Mark and Hanrahan, Pat},
  year={2005},
  school={Stanford university}
}

@article{adelson1992singlelenslightfield,
  title={Single lens stereo with a plenoptic camera},
  author={Adelson, EH and Wang, JYA},
  journal={IEEE Transactions on Pattern Analysis and Machine Intelligence},
  volume={14},
  number={2},
  pages={99--106},
  year={1992}
}

@ARTICLE{lightfieldsurvey2016,
  author={Ihrke, Ivo and Restrepo, John and Mignard-Debise, Lois},
  journal={IEEE Signal Processing Magazine}, 
  title={Principles of Light Field Imaging: Briefly revisiting 25 years of research}, 
  year={2016},
  volume={33},
  number={5},
  pages={59-69},
}

@ARTICLE{lightfieldsurvey2017,
  author={Wu, Gaochang and Masia, Belen and Jarabo, Adrian and Zhang, Yuchen and Wang, Liangyong and Dai, Qionghai and Chai, Tianyou and Liu, Yebin},
  journal={IEEE Journal of Selected Topics in Signal Processing}, 
  title={Light Field Image Processing: An Overview}, 
  year={2017},
  volume={11},
  number={7},
  pages={926-954},}

@inproceedings{caliskan2014tmwithopticalflow,
  title={Atmospheric turbulence mitigation using optical flow},
  author={Caliskan, Tufan and Arica, Nafiz},
  booktitle={2014 22nd International Conference on Pattern Recognition},
  pages={883--888},
  year={2014},
  organization={Ieee}
}

@INPROCEEDINGS{2008tmspline,
  author={Masao Shimizu and Shin Yoshimura and Masayuki Tanaka and Masatoshi Okutomi},
  booktitle={2008 IEEE Conference on Computer Vision and Pattern Recognition}, 
  title={Super-resolution from image sequence under influence of hot-air optical turbulence}, 
  year={2008},
  volume={},
  number={},
  pages={1-8}}

@inproceedings{ng2006lfaberration,
  title={Digital correction of lens aberrations in light field photography},
  author={Ng, Ren and Hanrahan, Patrick M},
  booktitle={International Optical Design Conference},
  pages={WB2},
  year={2006},
  organization={Optica Publishing Group}
}

@article{Wu2016lfthruturbulece,
author = {Chensheng Wu and Jonathan Ko and Christopher C. Davis},
journal = {Opt. Express},
number = {11},
pages = {11975--11986},
publisher = {Optica Publishing Group},
title = {Imaging through strong turbulence with a light field approach},
volume = {24},
month = {May},
year = {2016},
}

@article{loktev2011specklelfturbulence,
  title={Speckle imaging through turbulent atmosphere based on adaptable pupil segmentation},
  author={Loktev, Mikhail and Soloviev, Oleg and Savenko, Svyatoslav and Vdovin, Gleb},
  journal={Optics letters},
  volume={36},
  number={14},
  pages={2656--2658},
  year={2011},
  publisher={Optical Society of America}
}

@article{wu2019comparisonlfturbulence,
  title={Comparison between the plenoptic sensor and the light field camera in restoring images through turbulence},
  author={Wu, Chensheng and Paulson, Daniel A and Rzasa, John R and Davis, Christopher C},
  journal={OSA Continuum},
  volume={2},
  number={9},
  pages={2511--2525},
  year={2019},
  publisher={Optical Society of America}
}

@article{yoshida2017spectralnorm,
  title={Spectral norm regularization for improving the generalizability of deep learning},
  author={Yoshida, Yuichi and Miyato, Takeru},
  journal={arXiv preprint arXiv:1705.10941},
  year={2017}
}

@article{guo2024direct,
  title={Direct observation of atmospheric turbulence with a video-rate wide-field wavefront sensor},
  author={Guo, Yuduo and Hao, Yuhan and Wan, Sen and Zhang, Hao and Zhu, Laiyu and Zhang, Yi and Wu, Jiamin and Dai, Qionghai and Fang, Lu},
  journal={Nature Photonics},
  volume={18},
  number={9},
  pages={935--943},
  year={2024},
  publisher={Nature Publishing Group UK London}
}

@article{guo2024eventlfm,
  title={Event{LFM}: event camera integrated Fourier light field microscopy for ultrafast 3D imaging},
  author={Guo, Ruipeng and Yang, Qianwan and Chang, Andrew S and Hu, Guorong and Greene, Joseph and Gabel, Christopher V and You, Sixian and Tian, Lei},
  journal={Light: Science \& Applications},
  volume={13},
  number={1},
  pages={144},
  year={2024},
  publisher={Nature Publishing Group UK London}
}

@article{aguilar2014itf,
  title={Real-time video stabilization without phantom movements for micro aerial vehicles},
  author={Aguilar, Wilbert G and Angulo, Cecilio},
  journal={EURASIP Journal on Image and Video Processing},
  volume={2014},
  number={1},
  pages={46},
  year={2014},
  publisher={Springer}
}

@inproceedings{Lai-ECCV-2018-ewarp,
    author    = {Lai, Wei-Sheng and Huang, Jia-Bin and Wang, Oliver and Shechtman, Eli and Yumer, Ersin and Yang, Ming-Hsuan}, 
    title     = {Learning Blind Video Temporal Consistency}, 
    booktitle = {European Conference on Computer Vision},
    year      = {2018}
}

\newpage
\section{Supplementary Materials}




\begin{table}[h]
\caption{\textbf{Ablation of event-to-video methods.} E2VID methods trained on clean events fail to handle turbulence. Postprocessing with conventional turbulence mitigation also fails to recover scene quality, confirming the importance of our approach.}
\label{tab:e2vid}
\centering
\begin{tabular}{l|c|c|ccc}
\toprule
Method & Resolution & MambaTM Postprocess & PSNR $\uparrow$ & SSIM $\uparrow$ & LPIPS $\downarrow$ \\
\midrule
\multirow{4}{*}{E2VID+~\cite{e2vid+}} 
& $768 \times 768$ & \xmark & $14.51$ & $0.493$ & $0.5854$ \\
& $256 \times 256$ & \xmark & $14.51$ & $0.435$ & $0.5258$ \\
& $768 \times 768$ & \cmark & $14.48$ & $0.489$ & $0.5602$ \\
& $256 \times 256$ & \cmark & $14.48$ & $0.435$ & $0.5040$ \\
\midrule
\multirow{4}{*}{ET-Net~\cite{etnet}} 
& $768 \times 768$ & \xmark & $13.65$ & $0.485$ & $0.5952$ \\
& $256 \times 256$ & \xmark & $13.65$ & $0.417$ & $0.5383$ \\
& $768 \times 768$ & \cmark & $13.72$ & $0.478$ & $0.5744$ \\
& $256 \times 256$ & \cmark & $13.72$ & $0.414$ & $0.5215$ \\
\midrule
\multirow{4}{*}{HyperE2VID~\cite{ercan2024hypere2vid}} 
& $768 \times 768$ & \xmark & $12.83$ & $0.472$ & $0.6183$ \\
& $256 \times 256$ & \xmark & $12.83$ & $0.391$ & $0.5597$ \\
& $768 \times 768$ & \cmark & $12.96$ & $0.469$ & $0.5878$ \\
& $256 \times 256$ & \cmark & $12.97$ & $0.393$ & $0.5371$ \\
\midrule
\multirow{4}{*}{FireNet+~\cite{e2vid+}} 
& $768 \times 768$ & \xmark & $12.84$ & $0.402$ & $0.5982$ \\
& $256 \times 256$ & \xmark & $12.84$ & $0.352$ & $0.5502$ \\
& $768 \times 768$ & \cmark & $12.68$ & $0.392$ & $0.5872$ \\
& $256 \times 256$ & \cmark & $12.68$ & $0.344$ & $0.5451$ \\
\midrule
\multirow{3}{*}{Ours} 
& $768 \times 768$ & \xmark & $14.84$ & $0.501$ & $0.5185$ \\
& $256 \times 256$ & \xmark & $14.84$ & $0.461$ & $0.4607$ \\
& $3 \times 3 \times 256 \times 256$ & \xmark & $\mathbf{15.40}$ & $\mathbf{0.505}$ & $\mathbf{0.4332}$ \\
\bottomrule
\end{tabular}
\end{table}


\subsection{Ablation of Events to Video Methods}

\begin{figure}[t]
    \centering
    \includegraphics[width=\linewidth]{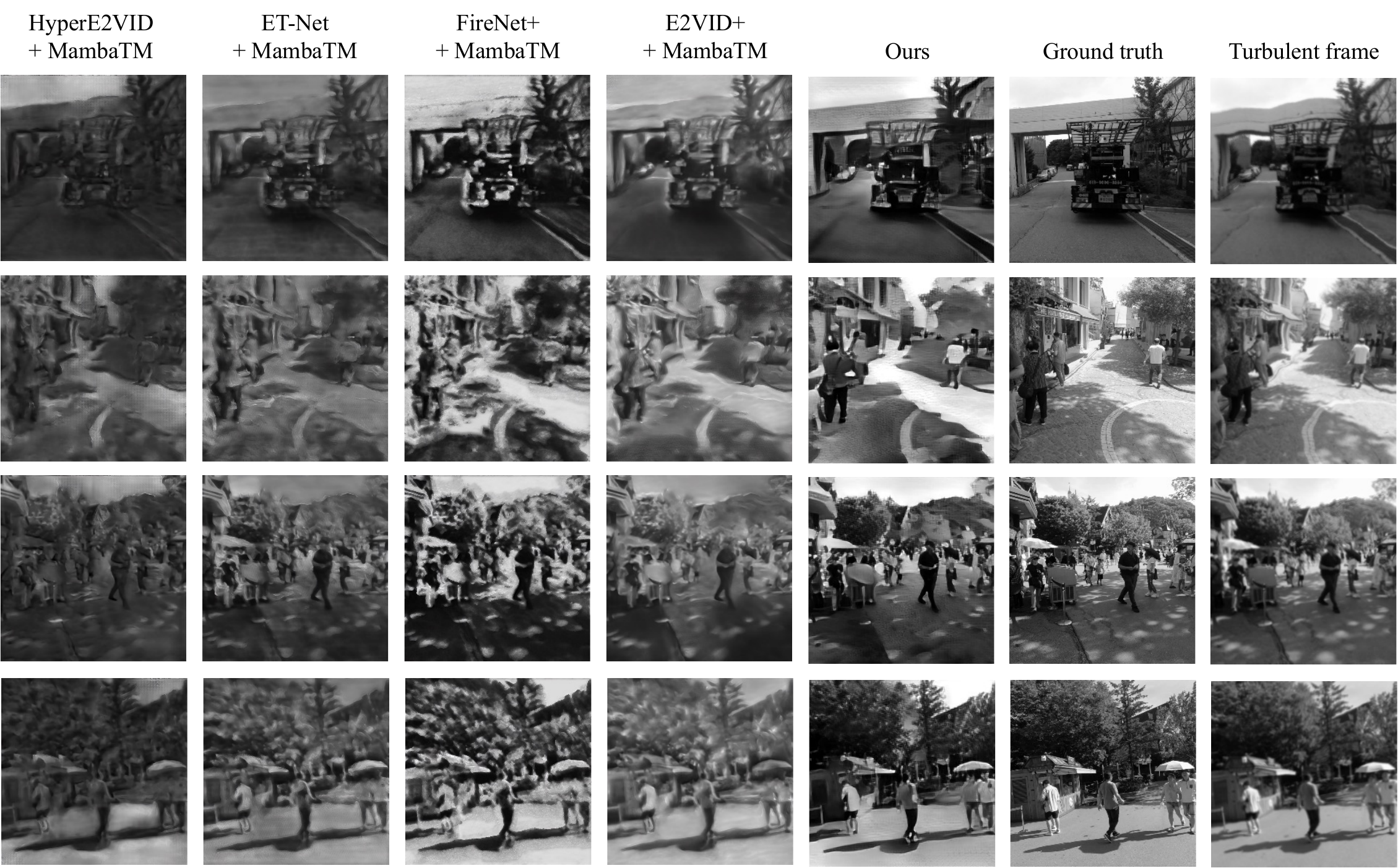}
    \caption{\textbf{Postprocessing event-to-video reconstructions with frame-based turbulence mitigation is insufficient.}
    Frames reconstructed by event-to-video methods and subsequently postprocessed by MambaTM~\cite{mambatm} retain significant geometric distortion and artifacts.
    The distortion arises because turbulence-induced events violate the fundamental assumption of event-to-video networks, producing corrupted reconstructions that MambaTM was not trained to handle.
    An end-to-end approach trained directly on turbulent event streams is necessary.}
    \label{fig:supp_e2vid}
\end{figure}


\begin{figure}[t]
    \centering
    \includegraphics[width=\linewidth]{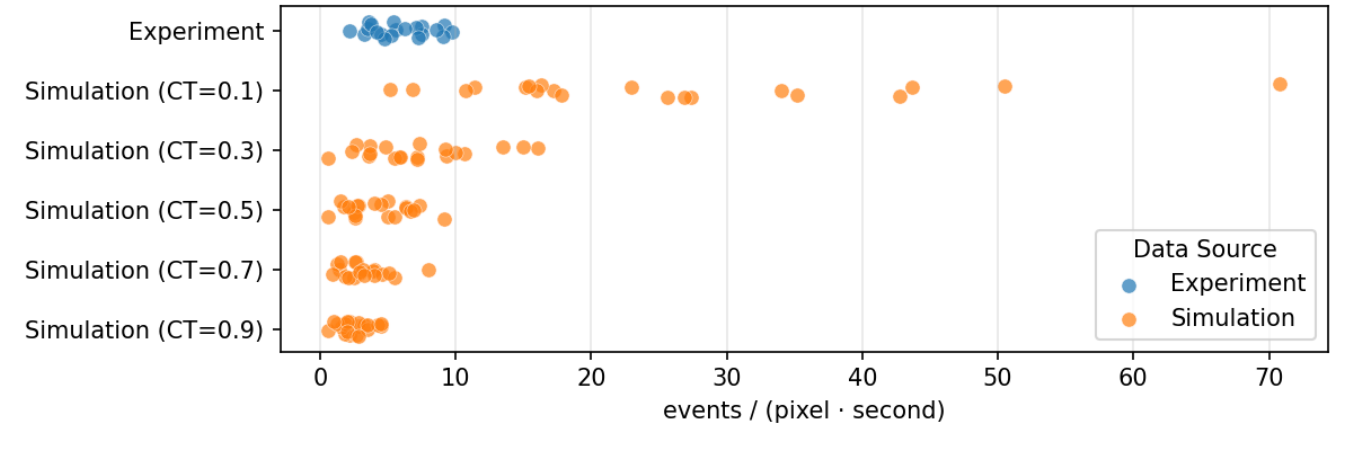}
    \caption{\textbf{Contrast threshold estimation for simulated dataset.} The average event rate (in $\text{events} \cdot \text{pix}^{-1} \cdot \text{s}^{-1}$) is computed from experimental light field event recordings and compared against synthetic light field event sequences generated with varying contrast threshold $C$. The experimental event rate falls within the range of synthetic rates for $C \in [0.1, 0.7]$. The range is deliberately picked wider than the experimental estimate to improve the network's robustness to contrast threshold variation across different scenes and lighting conditions.}
    \label{fig:ct}
\end{figure}

Existing event-to-video methods~\cite{e2vid, firenet, e2vid+} assume events arise exclusively from scene dynamics or camera motion. 
However, as argued in the introduction, dynamic turbulence produces events that are indistinguishable from actual scene motion. 
Consequently, these methods misinterpret turbulence as physical movement, irreparably entangling scene content with distortion in the reconstructed frames. 
This raises a natural follow-up question: could we simply reconstruct the video using these existing methods and then apply a state-of-the-art frame-based turbulence mitigation model, such as MambaTM~\cite{mambatm}, to clean the output? 
We demonstrate that this pipelined approach fails. 
Because methods like E2VID+, FireNet+, ET-Net, and HyperE2VID produce heavily corrupted reconstructions from turbulent streams, feeding their outputs into MambaTM yields no meaningful recovery. 
This proves that treating reconstruction and mitigation as separate, sequential steps is insufficient, and an end-to-end approach trained directly on turbulent event streams is necessary.

As illustrated in \autoref{tab:e2vid}, all event-to-video methods yield lower PSNR, SSIM, and LPIPS than our approach at equivalent pixel counts, confirming that turbulence-induced events violate the fundamental assumption these methods rely upon.
Postprocessing each reconstructed output with MambaTM yields negligible improvement, and in some cases degrades performance, demonstrating that frame-based turbulence mitigation cannot compensate for the distortion introduced upstream.

The qualitative results of postprocessing are showcased in \autoref{fig:supp_e2vid}.
The reconstructed frames retain significant geometric distortion, artifacts, and blur regardless of the event-to-video method used.
This outcome is expected as, on one hand, event-to-video networks were trained assuming events encode scene motion, so turbulence-induced events corrupt their hidden state and produce distorted reconstructions and even artifacts.
On the other hand, the resulting frames do not resemble the warp-plus-blur degradation that MambaTM was trained to correct, rendering the postprocessing ineffective.
All in all, these results confirm that our end-to-end approach trained directly on turbulent event streams is necessary for effective high-speed reconstruction through turbulence.

\subsection{Characterization of Contrast Threshold for Tabletop Setup}

The contrast threshold $C$ governs the sensitivity of an event camera to intensity changes and is a critical parameter for simulation-to-real transfer.
Stoffregen et al.~\cite{e2vid+} propose a proxy for estimating the experimental $C$ by tuning the simulator until the average event rate (measured in $\text{events} \cdot \text{pix}^{-1} \cdot \text{s}^{-1}$) matches that of real data.
Intuitively, higher $C$ reduces the event rate for a given scene, while faster motion increases it independently of $C$.

To estimate the appropriate $C$ for our setup, we compute the event rate from our experimental light field event recordings and compare it against synthetic light field event sequences generated with varying $C$ values.
The result is plotted in \autoref{fig:ct}, indicating that the experimental event rate falls within $C \in [0.1, 0.7]$.
We adopt this range for uniform sampling during training, with bounds intentionally wider than the observed experimental values to ensure robustness to contrast threshold variation across scenes and camera configurations.

\subsection{Characterization of Turbulence Strength for Tabletop Setup}
Here, the turbulence strength of our tabletop setup is characterized through three tilt statistics---temporal autocorrelation, spatial correlation within a view, and spatial correlation across views---together with the temporal evolution of blur and the scintillation effect. We also estimate the Fried parameter $r_0$ and refractive-index structure constant $C_n^2$.

\begin{figure}[t]
    \centering
    \includegraphics[width=\linewidth]{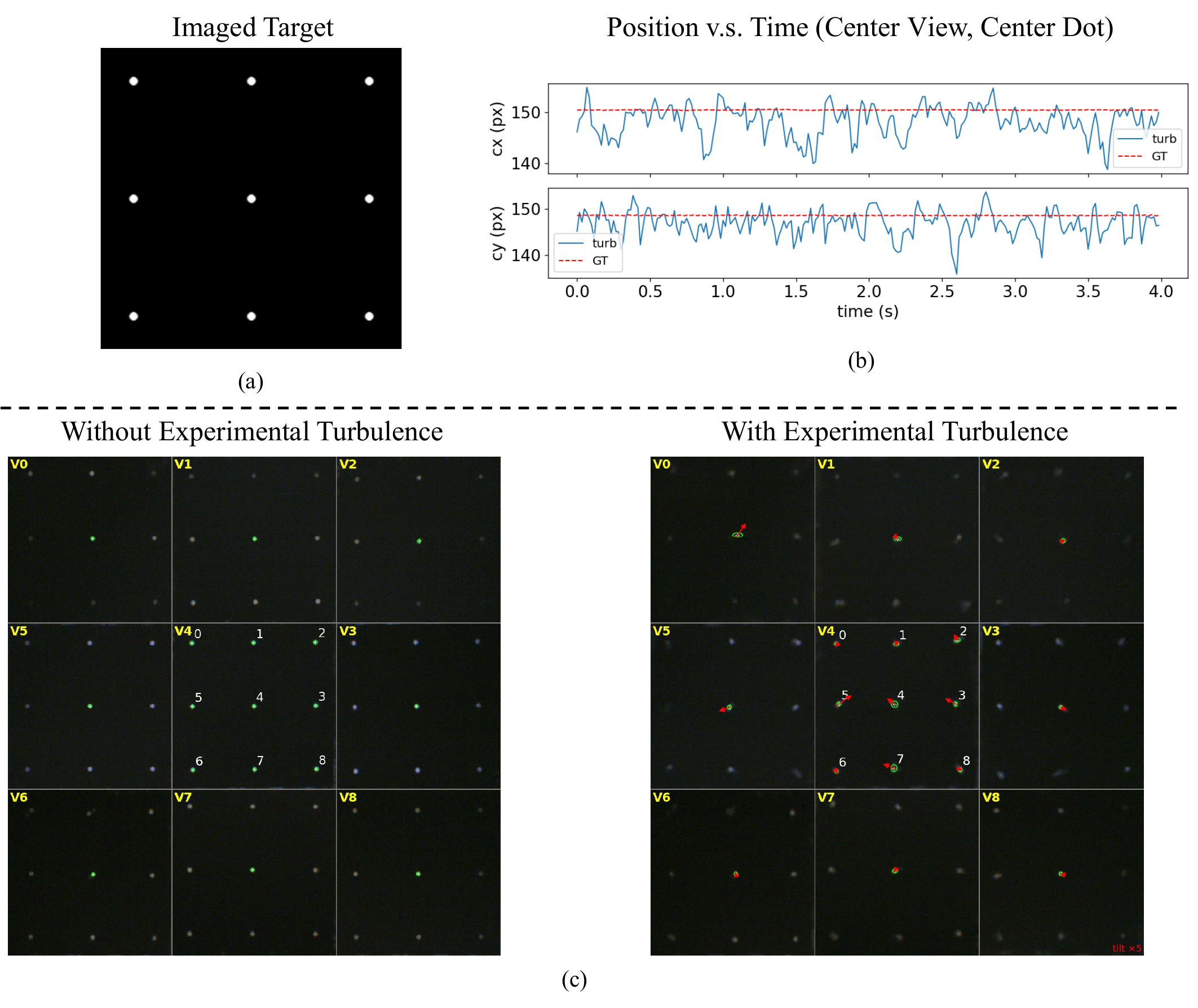}
    \caption{\textbf{Dot grid imaged through turbulence.} \textbf{(a)}~The $3\times3$ dot grid displayed on the monitor, used as the measurement target. \textbf{(b)}~Centroid position vs.\ time under turbulence (colored) and ground truth (GT, gray); the top and bottom rows show $x$- and $y$-displacement, respectively. \textbf{(c)}~Example frame illustrating turbulence-induced tilt (red arrows, magnified $5\times$ for visibility) and blur (green ellipses). Frames are recorded at 60\,fps with the video version shown in the webpage.}
    \label{fig:dot_grid}
\end{figure}

The measurement target is a $3 \times 3$ dot grid, shown in \autoref{fig:dot_grid}(a), imaged through the turbulent medium.
A 2D Gaussian is fitted to each dot to extract its centroid and spread; tilt and blur are then derived from these quantities as illustrated in \autoref{fig:dot_grid}(c).
As shown in \autoref{fig:dot_grid}(b), the centroid positions fluctuate substantially when the turbulence sources are active, whereas they remain stable in the ground-truth (GT) recording with turbulence off.

\begin{figure}[t]
    \centering
    \includegraphics[width=\linewidth]{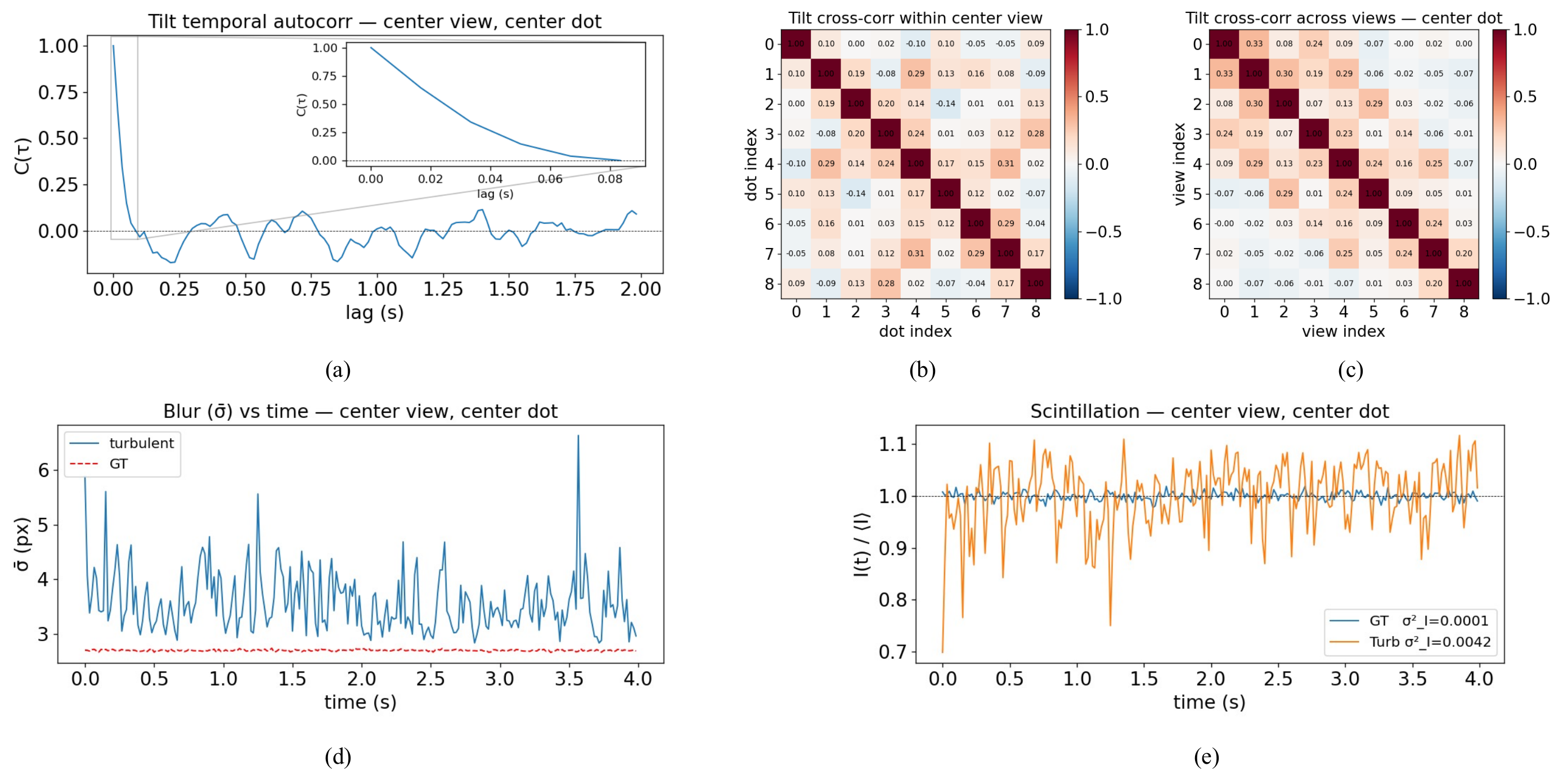}
    \caption{\textbf{Turbulence statistics of our tabletop setup.} We characterize turbulence-induced tilt and blur across time, space, and viewpoints. The $3\times3$ dots are indexed in row-alternating (snake) order. \textbf{(a)}~Tilt temporal autocorrelation decays within 0.08\,s, confirming the turbulence is dynamic. \textbf{(b)}~Nearby dots within the same view exhibit higher tilt correlation than distant ones, consistent with a finite spatial coherence length. \textbf{(c)}~Tilt across sub-apertures viewing the same scene location is largely uncorrelated, confirming that each viewpoint experiences distinct degradation. \textbf{(d)}~Blur $\bar{\sigma}$ fluctuates over time and is consistently larger than GT, confirming turbulence-induced PSF spreading. \textbf{(e)}~Normalized intensity $I(t)/\langle I\rangle$ for GT and turbulence recordings.
  The scintillation index $\sigma^2_I = \langle I^2\rangle/\langle I\rangle^2 - 1$ is larger under turbulence, confirming measurable intensity fluctuations induced
  by the tabletop setup.}
    \label{fig:turb_stats}
\end{figure}

For each of the 9 sub-aperture views extracted from the light field, a 2D Gaussian                                                                                                       

$$I(x,y) = A \exp\left(-\frac{(x-c_x)^2}{2\sigma_x^2} - \frac{(y-c_y)^2}{2\sigma_y^2}\right) + B,
$$
is fitted to each dot via least squares, yielding the centroid $(c_x, c_y)$ and the spread $(\sigma_x, \sigma_y)$. Tilt is defined as the deviation of the centroid from a no-turbulence reference, $\boldsymbol{\alpha} = (c_x - c_x^{\mathrm{ref}}, c_y - c_y^{\mathrm{ref}})$, where the reference centroid is obtained from the GT recording with the turbulence sources turned off. Blur is defined as the mean isotropic spread $\bar{\sigma} = (\sigma_x + \sigma_y)/2$. Statistics are computed over 240
consecutive frames spanning 4 seconds. 

The statistics are shown in \autoref{fig:turb_stats}. Normalized temporal autocorrelation is computed for the center dot of the center sub-aperture view. For tilt, the 2D vector autocorrelation at lag $\tau$ is

$$
C_{\mathrm{tilt}}(\tau) = \frac{1}{2}\left[\frac{\langle \alpha_x(t)\,\alpha_x(t+\tau)\rangle}{\langle \alpha_x^2 \rangle} + \frac{\langle \alpha_y(t)\,\alpha_y(t+\tau)\rangle}{\langle \alpha_y^2 \rangle}\right],
$$
which is equivalent to $\langle \boldsymbol{\alpha}(t)\cdot\boldsymbol{\alpha}(t+\tau)\rangle / \langle|\boldsymbol{\alpha}|^2\rangle$ when the system is isotropic ($\langle\alpha_x^2\rangle = \langle\alpha_y^2\rangle$).
As shown in \autoref{fig:turb_stats}(a), the tilt autocorrelation decays to zero within $\tfrac{1}{12}$\,s, indicating that the turbulence is dynamic and rapidly changing.

Normalized spatial correlation within a view is computed over the $9$ dots of the center sub-aperture view. For tilt, the pairwise correlation between dots $i$ and $j$ is

$$
C_{\mathrm{tilt}}[i,j] = \frac{\langle \boldsymbol{\alpha}_i(t)\cdot\boldsymbol{\alpha}_j(t)\rangle}{\sqrt{\langle|\boldsymbol{\alpha}_i|^2\rangle,\langle|\boldsymbol{\alpha}_j|^2\rangle}}.
$$
As shown in \autoref{fig:turb_stats}(b), the tilt spatial correlation shows a pattern where close dots are more correlated than far dots, consistent with the physical intuition that turbulence has a finite spatial coherence length. Overall, however, the spatial correlation is low, which is expected as the turbulence introduces spatially varying degradation.

Normalized spatial correlation across views is computed over the $9$ sub-aperture views for the center dot. 
The same vector dot-product correlation formula is applied. 
As shown in \autoref{fig:turb_stats}(c), the tilt spatial correlation across views is also low, which matches our insight that for each sub-aperture, the turbulence-induced degradation is different, as each sub-aperture samples a distinct ray bundle through the turbulent medium.

For blur, we plot $\bar{\sigma}(t)$ directly over time in \autoref{fig:turb_stats}(d), which shows clearly that the blur size fluctuates throughout the sequence, making dots approximately 1 to 2 times larger than in the ground truth. This further confirms the dynamic nature of the turbulence. 

 Scintillation characterizes intensity fluctuations caused by small temperature variations in the
atmosphere along turbulent propagation paths.
The scintillation index is defined as
  \begin{equation}
      \sigma^2_I \;=\; \frac{\langle I^2 \rangle}{\langle I \rangle^2} - 1
      \;=\; \frac{\mathrm{Var}(I)}{\langle I \rangle^2}.
      \label{eq:scint}
  \end{equation}
This quantity is dimensionless and equals zero for a perfectly stable beam.
As shown in \autoref{fig:turb_stats}(e), the turbulence recording exhibits a larger $\sigma^2_I$ than the GT recording, confirming that the tabletop setup introduces measurable intensity scintillation in addition to tilt and blur.

Finally, we estimate the Fried parameter from the measured PSF width using
$r_0 \approx 0.98 \frac{\lambda}{\mathrm{FWHM}}$,
where $\lambda$ is the wavelength and $\mathrm{FWHM}$ is the angular full width at half maximum of the turbulence-induced blur, derived from the fitted 2D Gaussian. We then estimate the refractive-index structure constant using
$C_n^2 = \frac{1}{0.423 k^2 L r_0^{5/3}}$, where
$k = \frac{2\pi}{\lambda}$.
This gives $r_0 = 0.012$\,m and $C_n^2 = 3.2 \times 10^{-12}\,\mathrm{m}^{-2/3}$.

\begin{table}[h]
\caption{\textbf{View-count vs.\ resolution trade-off in simulation}, showing that greater view diversity improves video stability.}
\label{table:resolution}
\centering
\begin{tabular}{l|c|cccc|c|cccc}
\toprule
\multirow{2}{*}{View} & \multicolumn{5}{c|}{Native Resolution} & \multicolumn{5}{c}{Upsampled to Match SV} \\
\cmidrule(lr){2-6}\cmidrule(lr){7-11}
 & Res. & PSNR $\uparrow$ & LPIPS $\downarrow$ & ITF $\uparrow$ & $E_{\text{warp}}$ $\downarrow$ & Res. & PSNR $\uparrow$ & LPIPS $\downarrow$ & ITF $\uparrow$ & $E_{\text{warp}}$ $\downarrow$ \\
\midrule
SV          & $768$ & $14.84$ & $0.519$          & $22.71$          & $0.0050$          & $768$ & $14.84$          & $\mathbf{0.519}$ & $22.71$          & $0.0050$ \\
$3\times3$  & $256$ & $15.40$ & $\mathbf{0.433}$ & $25.16$          & $0.0022$          & $768$ & $\mathbf{15.19}$ & $0.544$          & $25.42$          & $0.0022$ \\
$5\times5$  & $160$ & $\mathbf{15.41}$ & $0.443$ & $\mathbf{29.85}$ & $\mathbf{0.0010}$ & $768$ & $15.14$          & $0.606$          & $\mathbf{30.11}$ & $\mathbf{0.0010}$ \\
\bottomrule
\end{tabular}
\end{table}

\subsection{Light Field Views and Resolution Tradeoff on Performance}
We fully characterize the light field versus resolution tradeoff with simulation evaluations for the $1\times1$, $3\times3$, and $5\times5$ fields supported by the kaleidoscope. 
We also add PSNR metrics that account for lost resolution by upscaling low-resolution results to $768\times768$ before score computation. 
The increased LPIPS scores after upsampling reflect the resolution trade-off of distributing the sampling budget across more views.
However, PSNR remains similar, while moving from single view to 3$\times$3 and 5$\times$5 views increases ITF from 22.71 to 25.16 and 29.85, while reducing warping error from 0.0050 to 0.0022 and 0.0010. 
In conclusion, the multi-view configurations achieve better temporal stability and comparable or higher PSNR. 

\subsection{Ablation Study}

\begin{figure}[t]
    \centering

    \begin{subfigure}[t]{0.49\linewidth}
        \centering
        \includegraphics[width=\linewidth]{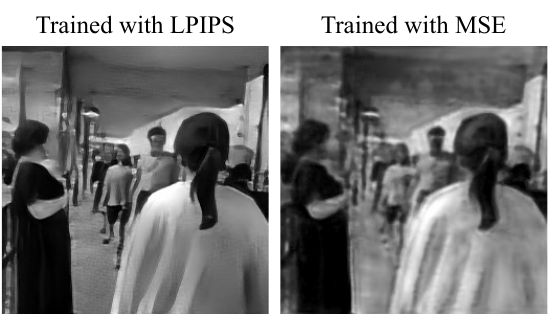}
        \caption{Replacing LPIPS with MSE.}
        \label{fig:ablate_mse}
    \end{subfigure}
    \hfill
    \begin{subfigure}[t]{0.49\linewidth}
        \centering
        \includegraphics[width=\linewidth]{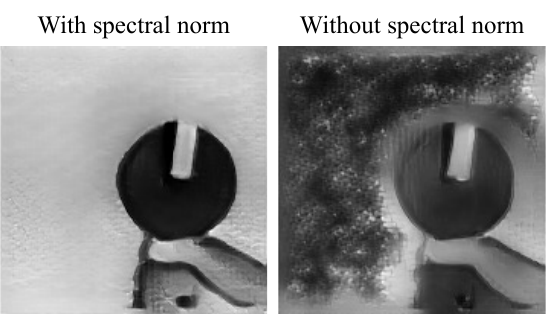}
        \caption{Removing spectral normalization.}
        \label{fig:ablate_spectral_norm}
    \end{subfigure}

    \caption{\textbf{Qualitative ablation of training objectives and regularization.}
    Compared with our LPIPS-trained baseline, replacing LPIPS with MSE introduces visible blur due to event-only intensity ambiguity.
    Removing spectral normalization produces severe artifacts in textureless regions.}
    \label{fig:ablation}
\end{figure}

We ablate the training objective and regularization to assess their effects.
Compared with our LPIPS-trained baseline (PSNR: 15.40, LPIPS: 0.433), replacing LPIPS with MSE introduces visible blur due to event-only intensity ambiguity, as shown in~\autoref{fig:ablate_mse}.
Although MSE increases PSNR to 16.08, it substantially degrades perceptual quality, increasing LPIPS to 0.579.
In addition, as illustrated in~\autoref{fig:ablate_spectral_norm}, removing spectral normalization produces severe artifacts in textureless regions, whereas spectral normalization yields a more stable background reconstruction. Additional ablations on pause augmentation and noise augmentation are reported in the original E2VID papers~\cite{e2vid,e2vid+}.

\subsection{ITF and Warping Error Calculation}
We use two metrics to quantify temporal consistency. ITF~\cite{aguilar2014itf} measures the average similarity between consecutive reconstructed frames:
\begin{equation}
\mathrm{ITF}(V) =
\frac{1}{T-1}
\sum_{t=1}^{T-1}
\mathrm{PSNR}(V_t,V_{t+1}),
\end{equation}
where $V_t$ is the reconstructed frame at time $t$ and $T$ is the total number of frames. Higher ITF indicates smaller frame-to-frame appearance changes.

We also report warping error following Lai et al.~\cite{Lai-ECCV-2018-ewarp}, which measures temporal consistency after compensating for motion. We estimate optical flow using the Farneback method, backwarp $V_{t+1}$ to the coordinate frame of $V_t$, and compute the valid-pixel mask using forward-backward flow consistency. For two consecutive frames, we compute
\begin{equation}
E_{\text{warp}}(V_t,V_{t+1}) =
\frac{1}{\sum_{j=1}^{N} M_t^{(j)}}
\sum_{j=1}^{N} M_t^{(j)}
\left\| V_t^{(j)} - \hat{V}_{t+1\rightarrow t}^{(j)} \right\|_2^2,
\end{equation}
where $N$ is the number of pixels, $\hat{V}_{t+1\rightarrow t}$ is the backward-warped frame, and $M_t \in \{0,1\}^N$ is the valid-pixel mask that excludes occluded or out-of-bound pixels. The sequence-level warping error is the average over adjacent frame pairs:
\begin{equation}
E_{\text{warp}}(V) =
\frac{1}{T-1}
\sum_{t=1}^{T-1}
E_{\text{warp}}(V_t,V_{t+1}).
\end{equation}
Lower warping error indicates stronger temporal consistency after motion compensation.

\end{document}